\documentclass[referee,sn-vancouver,Numbered]{sn-jnl}

\renewenvironment{table}[1][]%
{\tableorg[#1]%
\tablebodyfont%
\renewcommand\footnotetext[2][]{{\removelastskip\vskip3pt%
\let\tablebodyfont\tablefootnotefont%
\hskip0pt\if!##1!\else{\smash{$^{##1}$}}\fi##2\par}}%
}{\endtableorg}

\usepackage[utf8]{inputenc}
\usepackage{graphicx}
\usepackage[ruled]{algorithm2e}
\usepackage{hyperref}
\usepackage{verbatim}
\usepackage{amsmath,amssymb,amsfonts}
\usepackage{amsthm}
\usepackage{mathrsfs}
\usepackage[title]{appendix}
\usepackage{xcolor}
\usepackage{textcomp}
\usepackage{manyfoot}
\usepackage{booktabs}
\usepackage{listings}
\usepackage{rotating}
\usepackage{array}
\usepackage{multirow}
\usepackage{acro}
\usepackage{tabularx}
\usepackage{float}
\usepackage{pdflscape}

\newcommand{\fref}[1]{\footnote{\href{#1}{#1}}}

\title{Diagnosis extraction from unstructured Dutch echocardiogram reports using span- and document-level characteristic classification}

\author*[1]{\fnm{Bauke} \sur{Arends}} \email{b.k.o.arends-4@umcutrecht.nl}
\author[1]{\fnm{Melle} \sur{Vessies}}
\author[1]{\fnm{Dirk} \spfx{van} \sur{Osch}}
\author[1]{\fnm{Arco} \sur{Teske}}
\author[1]{\fnm{Pim} \spfx{van der} \sur{Harst}}
\author[1]{\fnm{René} \spfx{van} \sur{Es}}
\author[2]{\fnm{Bram} \spfx{van} \sur{Es}}

\affil[1]{\orgdiv{Department of Cardiology}, \orgname{University Medical Center Utrecht}, \orgaddress{\city{Utrecht}, \country{The Netherlands}}}
\affil[2]{\orgdiv{Central Diagnostic Laboratory}, \orgname{University Medical Center Utrecht}, \orgaddress{\city{Utrecht}, \country{The Netherlands}}}

\DeclareAcronym{nlp}{
  short=NLP,
  long=natural language processing,
}
\DeclareAcronym{ehr}{
  short=EHR,
  long=electronic health record,
}
\DeclareAcronym{ner}{
  short=NER,
  long=named entity recognition,
}
\DeclareAcronym{all}{
  short=ALL,
  long=approximate list lookup,
}
\DeclareAcronym{cnn}{
  short=CNN,
  long=convolutional neural network,
}
\DeclareAcronym{umcu}{
  short=UMCU,
  long=University Medical Center Utrecht,
}
\DeclareAcronym{bow}{
  short=BOW,
  long=bag-of-words,
}
\DeclareAcronym{icd}{
  short=ICD,
  long=International Classification of Disease,
}
\DeclareAcronym{crf}{
  short=CRF,
  long=conditional random fields,
}
\DeclareAcronym{rnn}{
  short=RNN,
  long=recurrent neural network,
}
\DeclareAcronym{svm}{
  short=SVM,
  long=support vector machine,
}
\DeclareAcronym{lstm}{
  short=LSTM,
  long=long short-term memory,
}
\DeclareAcronym{bilstm}{
  short=biLSTM,
  long=bilateral LSTM,
}
\DeclareAcronym{tf-idf}{
  short=TF-IDF,
  long=term frequency-inverse document frequency,
}
\DeclareAcronym{gru}{
  short=GRU,
  long=gated recurrent unit,
}
\DeclareAcronym{qrnn}{
  short=QRNN,
  long=quasi-recurrent neural network,
}
\DeclareAcronym{bigru}{
  short=biGRU,
  long=bidirectional GRU,
}

\begin{document}

\abstract{Clinical machine learning research and AI driven clinical decision support models rely on clinically accurate labels. Manually extracting these labels with the help of clinical specialists is often time-consuming and expensive. This study tests the feasibility of automatic span- and document-level diagnosis extraction from unstructured Dutch echocardiogram reports. 
We included 115,692 unstructured echocardiogram reports from the UMCU a large university hospital in the Netherlands. A randomly selected subset was manually annotated for the occurrence and severity of eleven commonly described cardiac characteristics. We developed and tested several automatic labelling techniques at both span and document levels, using weighted and macro F1-score, precision, and recall for performance evaluation. We compared the performance of span labelling against document labelling methods, which included both direct document classifiers and indirect document classifiers that rely on span classification results. 
The SpanCategorizer and MedRoBERTa$.$nl models outperformed all other span and document classifiers, respectively. The weighted F1-score varied between characteristics, ranging from 0.60 to 0.93 in SpanCategorizer and 0.96 to 0.98 in MedRoBERTa$.$nl. Direct document classification was superior to indirect document classification using span classifiers. SetFit achieved competitive document classification performance using only 10\% of the training data. Utilizing a reduced label set yielded near-perfect document classification results. 
We recommend using our published SpanCategorizer and MedRoBERTa$.$nl models for span- and document-level diagnosis extraction from Dutch echocardiography reports. For settings with limited training data, SetFit may be a promising alternative for document classification.
}


\maketitle

\section{Background}\label{Intro}
Unstructured \ac{ehr} data contains valuable information for a broad spectrum of clinical machine learning applications, including the creation of clinical decision support systems, semi-automated report writing, and cohort identification. The extraction of accurate clinical labels is essential to realize these applications. Relying solely on structured data for this purpose often yields disappointing outcomes, primarily due to two key reasons. Firstly, collecting structured data has only recently gained momentum in clinical practice, leaving a large volume of historical data underutilized. Secondly, the structured data that is collected, may suffer from a lack of precision and reliability \cite{misset2008}. \Ac{icd} coding specifically was identified as unreliable for phenotyping \acp{ehr} \cite{DeHond2022, Anderson2015, Perlis2012}. Data annotation is identified as one of the main obstacles in developing clinical \ac{nlp} applications \cite{Spasic2020}. 
Therefore, utilizing labels extracted from unstructured data has the potential to enhance both data volume and data quality.
\\ \\
Echocardiography, the most commonly performed cardiac imaging diagnostic \cite{pearlman2007}, provides a detailed anatomical and functional description of a wide range of cardiac structures. Data from echocardiography reports are consequently used in many aspects of patient care, as well as many clinical trials. However, the heterogeneous format of the reports, as well as medical text characteristics such as abundant shorthand, domain-specific vocabularies, implicitly assumed knowledge, and spelling and grammar mistakes, make extracting accurate labels challenging. For label extraction, we often resort to automated techniques, because manual extraction by domain experts is both costly and time-consuming. 
\\ \\
Previous work on data extraction from echocardiography reports has primarily focused on extracting quantitative measurement values from structured, semi-structured and unstructured parts of the report using rule-based methods \cite{nath2016a, szeker2023, kaspar2019, patterson2017}. Rule-based text-mining systems such as MedTagger \cite{Liu2013}, Komenti \cite{Slater2020}, and cTAKES \cite{Savova2010} are examples of low-code tools that allow clinicians to develop and apply rules for rule-based text mining. These rule-based methods offer several advantages, as they are transparent, easily modifiable, and do not require large amounts of labelled training data. Furthermore, they can be quite effective despite their simplicity. While their performance can vary based on the developer's expertise and attention to detail, a more specific downside of rule-based methods is their inherent inability to generalize beyond the set of predefined rules.
\\ \\
\Ac{nlp} methods based on machine learning may overcome some of these disadvantages, as they are able to learn rules implicitly from labelled data. In the biomedical field, several open-source systems, such as GATE \cite{cunningham_getting_2013} and cTAKES \cite{Savova2010} are available to employ these methods. Additionally, an abundance of model architectures is available for label extraction, including token classification models \cite{saha2009}, \ac{crf} \cite{ruan2018}, \ac{rnn} such as \ac{lstm} \cite{qin2020} and transformers such as BERT \cite{richter-pechanski2021}, \ac{svm} \cite{Wang2012_2} and AutoML methods \cite{Mustafa2021}. However, in the broader field of \ac{ner} in medical imaging reports, there does not seem to be one overall best-performing method \cite{qin2020, sugimoto2021, garcia-largo2021}. For span identification performance in particular, multiple factors may influence performance, including model architecture and span characteristics such as span frequency, distinctive span boundaries and span length \cite{papay2020}.
\\ \\
\Ac{ner} in the medical imaging report domain has mostly been described in English texts \cite{nath2016a, patterson2017, navarro2023}. There are limited studies in other languages, such as Dutch \cite{puts2023, vanes2022}, German \cite{jantscher2023}, and Spanish \cite{ahumada2024}. Few publicly available pre-trained Dutch language models exist, and include BERTje \cite{devries2019} and
RobBERT \cite{Delobelle2020, delobelle2024}. Verkijk and Vossen recently created MedRoBERTa.nl, a version of RoBERTa \cite{liu2019} finetuned on Dutch \ac{ehr} data \cite{verkijk2021}. Furthermore, Remy, Demuynck and Demeester developed a multilingual large language model BioLORD-2023M using contrastive learning, which is able to identify biomedical concepts and sentences \cite{remy2024}. To the best of our knowledge, none of these models have been finetuned with the goal of information extraction from Dutch echocardiogram reports.
\\ \\
In this work, we focus on span and document label extraction from unstructured Dutch echocardiogram reports for a wide range of clinical concepts. To capture the most meaningful clinical concepts, we constructed a custom ontology which incorporates most major cardiac abnormalities. We explicitly focused on extracting qualitative labels from unstructured text, as several algorithms exist to extract measurement values from structured and semi-structured data. We evaluated three \ac{nlp} methods for span-level label extraction, and six \ac{nlp} methods for document-level label extraction. The best-performing span and document classification models are available on the Huggingface model repository\fref{https://huggingface.co/UMCUtrecht}. Additionally, the developed code is publicly available on GitHub\fref{https://github.com/umcu/EchoLabeler}.

\section{Methods}
This section provides a detailed description of our data and the data annotation process, followed by an overview of our experiments. We employ several \ac{nlp} methods for extracting span and document-level labels from Dutch echocardiogram reports. Additional information on model parameters is detailed in Additional File 1.

\subsection{Data overview}
Our dataset consisted of $115,692$ unstructured echocardiogram reports collected during routine clinical care from 2003 to 2023, stored in the \ac{ehr} at \ac{umcu}, a large university hospital in the Netherlands. Over this period, there has not been a universal standard for report writing. Reports containing fewer than fifteen characters were excluded, as were reports with fewer than thirty characters that lacked any description of a medical concept. These reports often contained only the phrase \textit{"For the report, see the patient's chart"}. 

\subsection{Data annotation}

\begin{figure}[ht]
    \centering
    \includegraphics[width=0.75\linewidth]{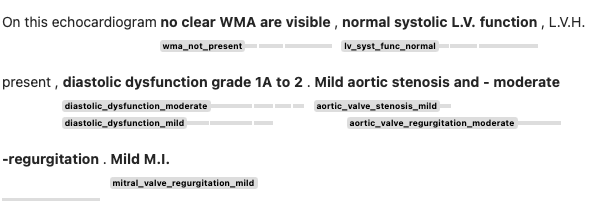}
    \caption{Example report with manual annotations. For presentation purposes, text has been translated to English.}
    \label{fig:example-report}
\end{figure}

In a randomly selected subset of the unstructured text portions of these reports, we manually annotated eleven common cardiac characteristics, which included left and right ventricular systolic function and dilatation, valvular disease, pericardial effusion, wall motion abnormalities, and diastolic dysfunction. Figure~\ref{fig:example-report} displays an example report including annotations. We assigned mutually exclusive labels for each characteristic to the span/document (Table~\ref{tab:label-def}).

\begin{table}[ht!]
\begin{tabular}{lr}
\hline
\textbf{Label} & \textbf{Description} \\
\hline
No label & No statement regarding this characteristic \\
Normal & Normal function described for this characteristic \\
Mild & Mildly abnormal function \\
Moderate & Moderately abnormal function \\
Severe & Severely abnormal function \\
Present & Abnormal function, unspecified severity \\ 
\bottomrule
\end{tabular}
\caption{Span and document label definitions}
\label{tab:label-def}
\end{table}

Annotations were checked sample wise by doctors. In cases of uncertainty, cases were jointly reviewed to achieve consensus. Several rounds of training iterations were completed before commencing the annotation task. To streamline the annotation process, each echocardiogram report was annotated for one characteristic at a time, resulting in eleven separate annotation files. For an overview of labelling instructions, see Additional file 2. Prodigy \cite{prodigy_montani_honnibal} was employed for the annotation task.

To ensure an adequate number of labels, we established the following requirements: for each characteristic, a minimum of 5000 documents were annotated, with the same documents used for each characteristic. In addition, to ensure sufficient training data, a minimum of 50 span labels per class, per characteristic were required, resulting in more than 5000 annotated documents for several characteristics (Table~\ref{tab:document-counts}). Document-level labels were constructed using the span-level labels. Given a multitude of span labels in one document, we aggregated the labels by selecting the most severe label per characteristic. For comparison we also employed a simplified label scheme with only three possible labels: not mentioned, normal, or present.

\begin{table}[ht]
\caption{Document label counts}
\label{tab:document-counts}
    \begin{tabular}{p{3cm}p{0.5cm}p{1cm}p{1cm}p{1cm}p{1cm}p{1cm}p{1cm}}
    \toprule
    Characteristic & Cases & Any label & Normal & Mild & Moderate & Severe & Present \\
    \midrule
    \raggedright Aortic regurgitation & 5615 & 2403 (42.8\%) & 1716 (30.6\%) & 505 (9.0\%) & 133 (2.4\%) & 49 (0.9\%) & 0 (0.0\%) \\
    \raggedright Aortic stenosis & 5000 & 1718 (34.4\%) & 1461 (29.2\%) & 108 (2.2\%) & 68 (1.4\%) & 81 (1.6\%) & 0 (0.0\%) \\
    \raggedright Diastolic dysfunction & 5000 & 1526 (30.5\%) & 521 (10.4\%) & 632 (12.6\%) & 243 (4.9\%) & 130 (2.6\%) & 0 (0.0\%) \\
    \raggedright Left ventricular dilatation & 5000 & 2402 (48.0\%) & 1870 (37.4\%) & 249 (5.0\%) & 91 (1.8\%) & 51 (1.0\%) & 141 (2.8\%) \\
    \raggedright Left ventricular systolic dysfunction & 5000 & 4503 (90.1\%) & 2881 (57.6\%) & 879 (17.6\%) & 378 (7.6\%) & 365 (7.3\%) & 0 (0.0\%) \\
    \raggedright Mitral regurgitation & 5000 & 2590 (51.8\%) & 1605 (32.1\%) & 733 (14.7\%) & 187 (3.7\%) & 65 (1.3\%) & 0 (0.0\%) \\
    \raggedright Pericardial effusion & 8686 & 1274 (14.7\%) & 973 (11.2\%) & 154 (1.8\%) & 55 (0.6\%) & 48 (0.6\%) & 44 (0.5\%) \\
    \raggedright Right ventricular dilatation & 8203 & 2718 (33.1\%) & 2137 (26.1\%) & 266 (3.2\%) & 125 (1.5\%) & 50 (0.6\%) & 140 (1.7\%) \\
    \raggedright Right ventricular systolic dysfunction & 5000 & 2462 (49.2\%) & 1807 (36.1\%) & 408 (8.2\%) & 188 (3.8\%) & 59 (1.2\%) & 0 (0.0\%) \\
    \raggedright Tricuspid regurgitation & 5000 & 1801 (36.0\%) & 1333 (26.7\%) & 262 (5.2\%) & 140 (2.8\%) & 66 (1.3\%) & 0 (0.0\%) \\
    \raggedright Wall motion abnormalities & 5000 & 1224 (24.5\%) & 389 (7.8\%) & 0 (0.0\%) & 0 (0.0\%) & 0 (0.0\%) & 835 (16.7\%) \\
    \bottomrule
    \end{tabular}
\end{table}

\begin{table}[ht]
\caption{Number of characteristics in each dataset}
\label{tab:corpus-table}
\begin{tabular}{lrr}
\toprule
Characteristic & Train & Test \\
\midrule
Aortic regurgitation & 2108 & 499 \\
Aortic stenosis & 1499 & 351 \\
Diastolic dysfunction & 1293 & 304 \\
Left ventricular dilatation & 2003 & 466 \\
Left ventricular systolic dysfunction & 4212 & 1035 \\
Mitral regurgitation & 2362 & 540 \\
Pericardial effusion & 1048 & 247 \\
Right ventricular dilatation & 2260 & 552 \\
Right ventricular systolic function & 2131 & 509 \\
Tricuspid regurgitation & 1574 & 380 \\
Wall motion abnormalities & 1075 & 259 \\
\bottomrule
\end{tabular}
\end{table}

\subsection{Data splits}

We split the dataset in a training and testing set, allocating $80$\% and $20$\%, respectively. We used one train/test split for a simple practical reason: we developed regular expressions for identifying candidate spans and for direct labelling only on the train split. Performing this extraction for $N$ folds would be infeasible and realistically also not truly independent unless we use different annotators for each fold. Since the amount of labelled cases may differ for each characteristic due to label prevalence and our requirements, a preemptive split was made using all $115,692$ reports. Consequently, the training and testing splits may not add up to exactly the prespecified percentages. In Table~\ref{tab:corpus-table}, we report the distribution of span-level labels for each data split.

\begin{table}[htbp]
\footnotesize
\caption{Span characteristics}
\label{tab:span_characteristics}
\begin{tabular}{lccccc}
\toprule
 &  & No. of spans & Length & SD & BD \\
Characteristic & Severity &  &  &  &  \\
\midrule
Aortic regurgitation & Overall & 2607 & 2.47 & 2.62 & 1.25 \\
 & Normal & 1849 & 2.48 & 2.42 & 1.28 \\
 & Mild & 562 & 2.39 & 3.08 & 1.05 \\
 & Moderate & 146 & 2.68 & 2.90 & 1.45 \\
 & Severe & 50 & 2.37 & 4.33 & 1.80 \\
\midrule
Aortic stenosis & Overall & 1850 & 2.48 & 2.60 & 1.35 \\
 & Normal & 1582 & 2.48 & 2.40 & 1.31 \\
 & Mild & 111 & 2.45 & 3.54 & 1.57 \\
 & Moderate & 73 & 2.53 & 3.43 & 1.71 \\
 & Severe & 84 & 2.39 & 4.33 & 1.52 \\
\midrule
Diastolic dysfunction & Overall & 1597 & 4.58 & 2.42 & 1.28 \\
 & Normal & 536 & 4.26 & 1.59 & 1.44 \\
 & Mild & 665 & 4.89 & 2.77 & 1.06 \\
 & Moderate & 263 & 4.73 & 2.92 & 1.41 \\
 & Severe & 133 & 3.96 & 3.01 & 1.46 \\
\midrule
Left ventricular dilatation & Overall & 2469 & 3.31 & 2.11 & 1.46 \\
 & Normal & 1925 & 3.42 & 1.80 & 1.30 \\
 & Mild & 256 & 3.21 & 2.98 & 1.88 \\
 & Moderate & 94 & 3.15 & 3.41 & 2.40 \\
 & Severe & 52 & 3.16 & 3.75 & 2.78 \\
 & Present & 142 & 2.19 & 3.35 & 1.75 \\
\midrule
Left ventricular systolic dysfunction & Overall & 5144 & 4.81 & 1.41 & 1.10 \\
 & Normal & 3113 & 4.85 & 1.28 & 1.01 \\
 & Mild & 1042 & 4.77 & 1.55 & 1.14 \\
 & Moderate & 495 & 4.82 & 1.53 & 1.28 \\
 & Severe & 494 & 4.64 & 1.82 & 1.39 \\
\midrule
Mitral regurgitation & Overall & 2902 & 2.56 & 2.56 & 1.33 \\
 & Normal & 1793 & 2.51 & 2.34 & 1.34 \\
 & Mild & 814 & 2.62 & 2.90 & 1.22 \\
 & Moderate & 228 & 2.74 & 2.82 & 1.46 \\
 & Severe & 67 & 2.71 & 3.31 & 1.69 \\
\midrule
Pericardial effusion & Overall & 1295 & 3.65 & 2.86 & 1.48 \\
 & Normal & 987 & 2.51 & 2.97 & 1.47 \\
 & Mild & 158 & 5.10 & 2.51 & 1.44 \\
 & Moderate & 55 & 11.50 & 2.15 & 1.70 \\
 & Severe & 50 & 12.81 & 2.21 & 1.65 \\
 & Present & 45 & 3.94 & 3.26 & 1.47 \\
\midrule
Right ventricular dilatation & Overall & 2812 & 3.54 & 2.06 & 1.40 \\
 & Normal & 2195 & 3.63 & 1.79 & 1.30 \\
 & Mild & 294 & 3.43 & 2.88 & 1.72 \\
 & Moderate & 132 & 3.30 & 3.28 & 1.91 \\
 & Severe & 50 & 3.43 & 3.52 & 1.97 \\
 & Present & 141 & 2.51 & 2.89 & 1.64 \\
\midrule
Right ventricular systolic dysfunction & Overall & 2640 & 4.34 & 1.70 & 1.37 \\
 & Normal & 1932 & 4.29 & 1.65 & 1.37 \\
 & Mild & 445 & 4.63 & 1.74 & 1.31 \\
 & Moderate & 199 & 3.98 & 2.05 & 1.44 \\
 & Severe & 64 & 5.11 & 2.07 & 1.51 \\
\midrule
Tricuspid regurgitation & Overall & 1954 & 2.47 & 2.84 & 1.48 \\
 & Normal & 1422 & 2.48 & 2.57 & 1.49 \\
 & Mild & 294 & 2.46 & 3.34 & 1.40 \\
 & Moderate & 165 & 2.37 & 3.72 & 1.50 \\
 & Severe & 73 & 2.53 & 4.01 & 1.56 \\
\midrule
Wall motion abnormalities & Overall & 1334 & 3.81 & 2.33 & 1.10 \\
 & Normal & 421 & 3.42 & 2.38 & 1.09 \\
 & Present & 913 & 4.00 & 2.31 & 1.10 \\
 \bottomrule
\end{tabular}
\footnotetext{\textit{Abbreviations.} SD, span distinctiveness; BD, span boundary distinctiveness.}
\end{table}

\subsection{Span classification}

We present three approaches for span classification. First, we employed a rule-based approach using regular expressions as a baseline method. Second, we used a \ac{ner}+L extractor, where clinical concept spans are identified and subsequently classified. Finally, we implemented a greedy span classification approach, where all possible spans are classified, and only those exceeding a threshold model certainty are presented.

\subsubsection{Approximate list lookup}

Given a dictionary of lists containing phrases, where each list represents a target label, we can build a very simple pseudo-model. This model indexes the phrases using phrase embeddings, then uses these embeddings to convert spans from unseen texts into fuzzy search keys. We used a dictionary with token-based regular expressions to extract matching phrases, denoting this method as \ac{all}. The advantage of this approach lies in its transparency and the ease with which the pseudo-model can be improved by simply adding or removing phrases. The rule-based algorithm was constructed as follows:  

\begin{algorithm}[H]
 \KwData{Document $\mathbf{d}$}
 \KwResult{Dictionary with labels}
 \For{span in spans}{
     \For{label in labels}{
        \If{PhraseMatcher(span, label) is True}{
            return label\;
        }
    }
 }
\caption{Rule-based look-up algorithm (ALL$_{rule}$)}
\end{algorithm}

\subsubsection{MedCAT}

\begin{figure}[ht]
    \centering
    \includegraphics[width=0.75\linewidth]{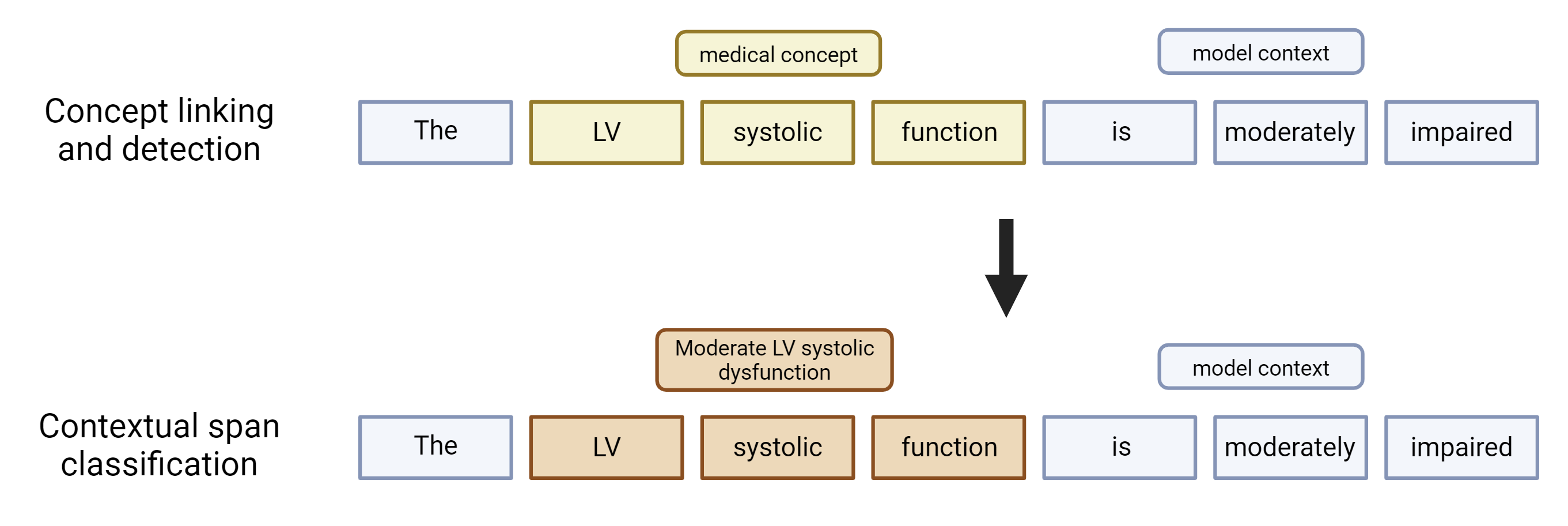}
    \caption{MedCAT pipeline for identifying and classifying medical concepts}
    \label{fig:medcat-example}
\end{figure}

MedCAT is a semi-supervised \ac{ner}+L extractor that supports \ac{bilstm} and transformer-based span-classifiers \cite{Kraljevic2021} (Figure~\ref{fig:medcat-example}). The benefit is that not all token spans are scanned. However, this requires training the MedCAT model to create a context-database that contains context vectors that are indicative for medical concepts. We performed unsupervised training on the training split and added the spans that were defined during the manual labelling process to MedCAT's vocabulary and context-database. The initial span-detector introduces a selection bias compared to a greedy span-classifier. Consequently, we expected a higher precision but lower recall, as some spans may be missed. We trained a different span classifier for each characteristic where all classifiers were integrated into one MedCAT modelpack. To reduce the occurrence of false negatives, we explicitly added a negative label for each class, set to $1$ whenever a class was otherwise unlabelled.

\subsubsection{spaCy SpanCategorizer}

\begin{figure}[ht]
    \centering
    \includegraphics[width=0.75\linewidth]{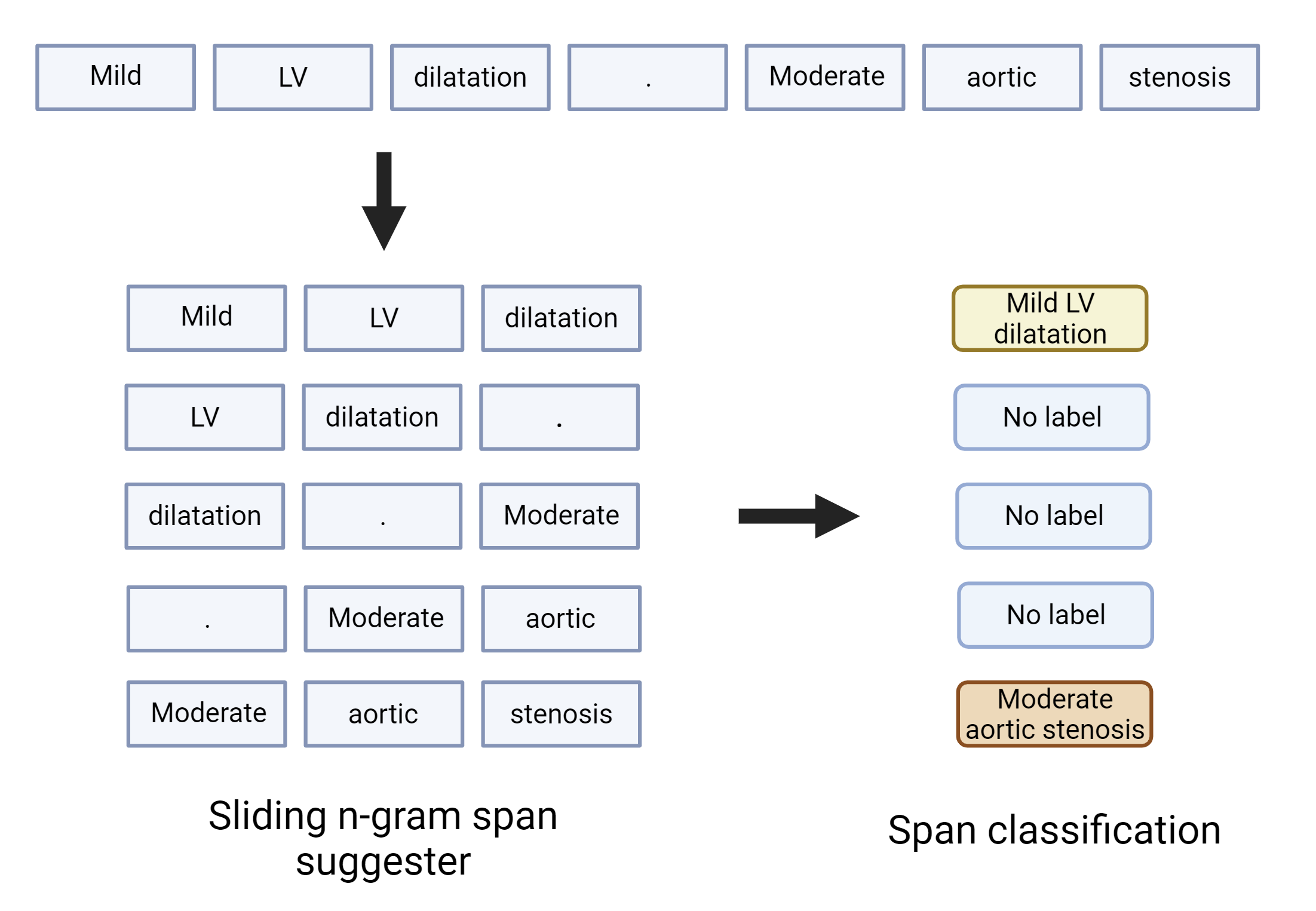}
    \caption{SpanCat pipeline for iterating and classifying n-gram spans using scanning windows of $1$-$25$ tokens}
    \label{fig:spancat-example}
\end{figure}

Similar to MedCAT, spaCy's SpanCategorizer \cite{honnibal2020spacy} operated in two stages: tokenization and span suggestion, followed by span classification. It employed a rule-based Dutch text tokenizer from spaCy. Unlike MedCAT, SpanCategorizer's default span suggester is greedy, suggesting all n-gram spans within a prespecified range of span lengths (Figure~\ref{fig:spancat-example}). The range for the n-gram suggester was set at $1$-$25$, due to the expected lengthy sentences describing some of the characteristics. Compared to a pipeline with a stricter span suggester, this setup was expected to yield a higher end-to-end recall but a corresponding lower precision due to an increase in false positives.
\\ \\
The span classification pipeline involved several steps. Tokens from suggested spans were first embedded using a multi hash embedding function based on a token's lexical attributes, followed by encoding using max-out activation, layer normalization, and residual connections. These encoded representations underwent mean and max pooling before being passed through a hidden layer. Finally, single label classification was performed on these span vectors using a logistic loss function. Each span was classified into one of the labelled classes, or was classified with a negative label (i.e., no label). 
\\ \\
Standard components and configuration files were predominantly used to prevent overfitting. However, for some characteristics, $>$$70$\% of cases had only negative labels (Table~\ref{tab:span_characteristics}). To address this imbalance and to prevent our model from solely predicting negative labels, different weights were assigned to negative labels ($0.6$, $0.8$ and $1.0$). For each characteristic, models were trained with these weights, and the model yielding the highest weighted F1-score was selected.

\subsection{Document classification}

For document classification, we used six methods. We implemented two baseline methods: one utilizing a \ac{bow} approach with medical word embeddings, and the one using indirect document classification via a span-to-document label heuristic, where the best performing span classification method was used to aggregate span-based classifications into document classifications. We also employed SetFit in combination with the multilingual BioLORD-2023 embeddings. Another method involved using RoBERTa, specifically the MedRoBERTa.nl model for this work. Additionally, we applied a \ac{rnn} model, specifically a bidirectional GRU, and a bidirectional \ac{cnn}.

\subsubsection{Bag-of-words}

\begin{figure}
    \centering
    \includegraphics[width=0.75\linewidth]{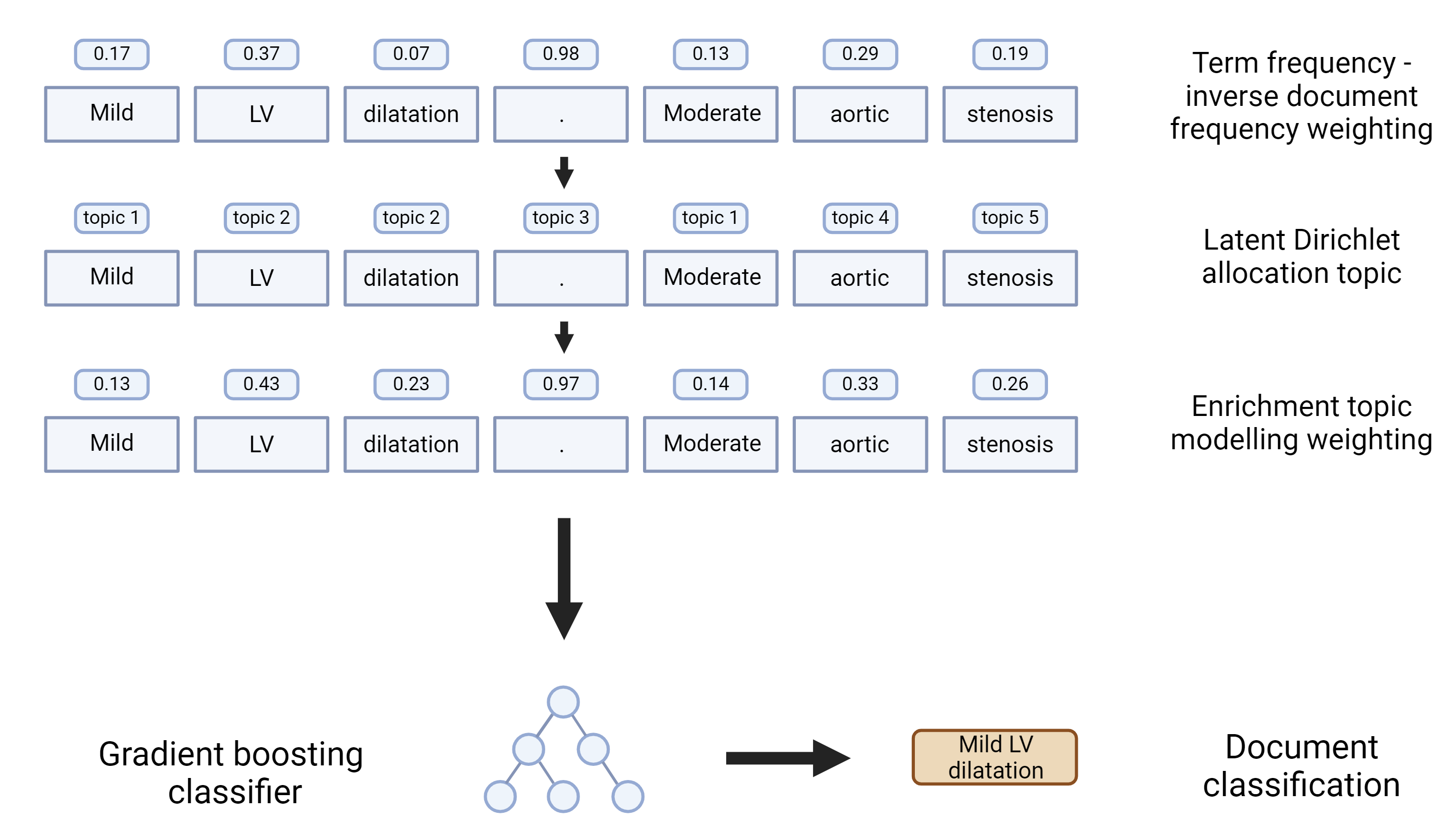}
    \caption{BOW pipeline involving tokenization, \ac{tf-idf} weighting, topic modelling, and classification using a gradient-boosted classifier}
    \label{fig:bow-pipeline}
\end{figure}

Our baseline \ac{bow} approach involved several feature extraction steps, detailed in Figure~\ref{fig:bow-pipeline}. First, the text was tokenized. Next, we applied \ac{tf-idf} weighting to each token within a document. We then enriched the features with topic modelling weighting, as described by Bagheri et al. \cite{Bagheri2020}. Additionally, we augmented the features with latent Dirichlet allocation topic probabilities to capture underlying thematic structures. The resulting features were combined with a standard gradient-boosted classifier. 

\subsubsection{Span classifier heuristic}
We selected the best performing span classifier based on its end-to-end performance. Then, we aggregated the span labels into a document label for each characteristic. The process is similar to how we constructed document labels: given a multitude of span labels within one document, we aggregated them by selecting the most severe label per characteristic. This heuristic allowed for more granular analysis by indicating which spans lead to the document classification. However, we expected performance loss due to the increased complexity of span classification.

\subsubsection{SetFit}

\begin{figure}
    \centering
    \includegraphics[width=0.75\linewidth]{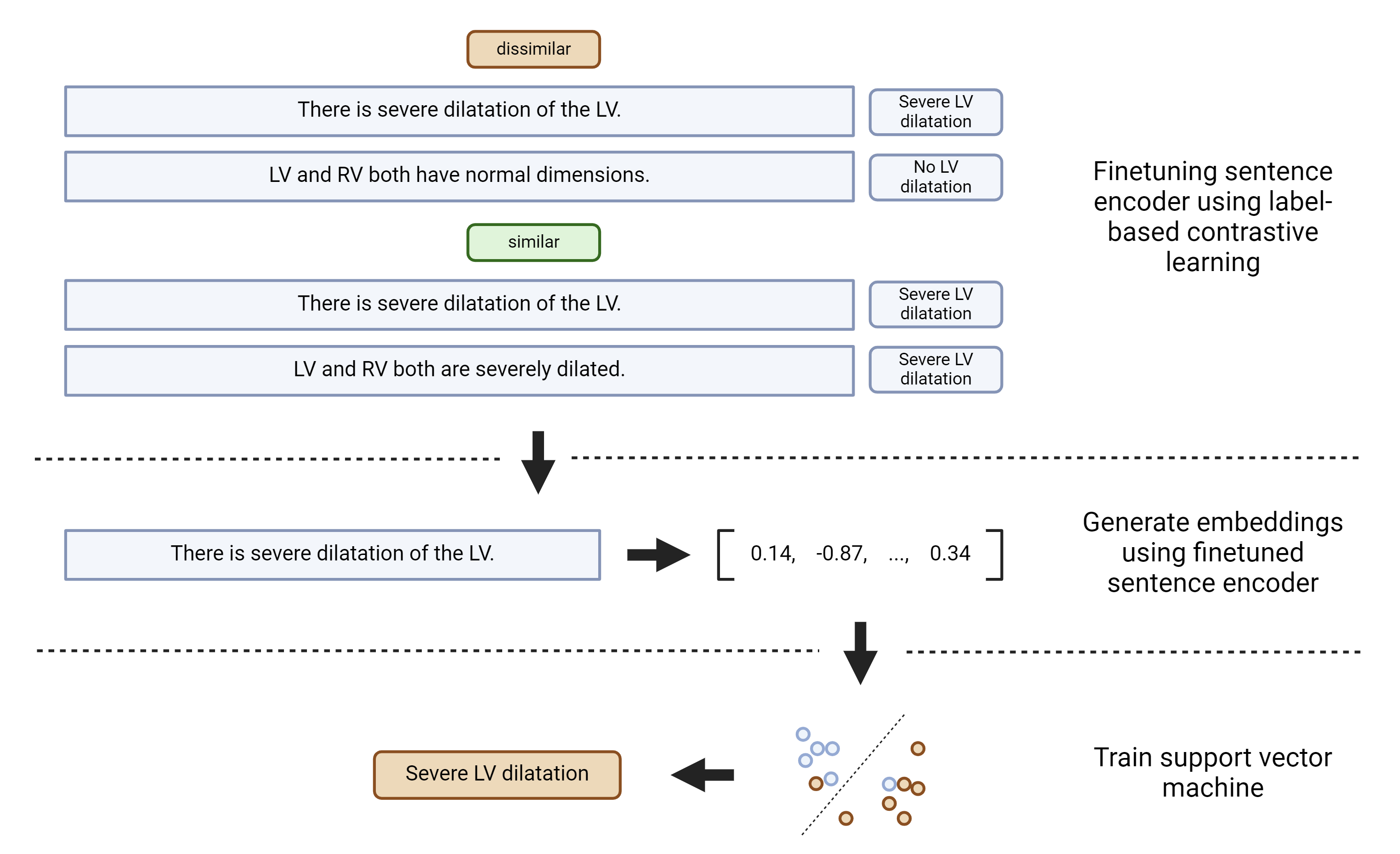}
    \caption{SetFit pipeline: fine-tuning the sentence encoder with label-based contrastive learning, followed by classification}
    \label{fig:setfit-pipeline}
\end{figure}

Reimers et al. \cite{Reimers2019} employed Siamese networks with contrastive learning on similar and dissimilar sentences to produce transformer-based encoders that capture semantic information along different axes of similarity, such as polarity and temporality. Tunstall et al. \cite{tunstall_efficient_2022} utilized these so-called "sentence encoders" to develop  SetFit, a few-shot classification method. SetFit fine-tunes the sentence encoder via contrastive learning based on labels, followed by training a classification head on the finetuned weights, as described in Figure~\ref{fig:setfit-pipeline}. \\ \\
For our work, we used BioLORD2023 developed by Remy et al. \cite{remy2024} and a sentence encoder that was trained on top of the RobBERTv2 model (see \cite{Delobelle2020,Delobelle2021}) by the Dutch Institute for Forensics. BioLORD-2023M serves as a multilingual sentence encoder designed to discriminate between medical concepts using existing ontologies. For the classification head we used a $\mu-$\ac{svm} model, a technique also applied by Beliveau et al. \cite{Beliveau2024}, who reported varying performance among state-of-the-art classification models. We trained the SetFit model using $500$ samples, which constitutes approximately $10\%$ of the total training dataset.

\subsubsection{MedRoBERTa.nl}

The BERT model, developed by Devlin et al. \cite{Devlin2019}, is a transformer-based machine learning technique known as Bidirectional Encoder Representations from Transformers. It is designed to pre-train deep bidirectional representations by jointly conditioning on both left and right context in all layers. This allows the model to be fine-tuned with just one additional output layer to create state-of-the-art models for a wide range of tasks, such as question answering and language inference, without substantial task-specific architecture modifications. BERT is built around the transformer architecture, which uses the self-attention mechanism (see Vaswani et al. \cite{Vaswani2017}) that weigh the influence of different words on each other irrespective of their position in the sentence.
\\ \\
RoBERTa, a robustly optimized version of BERT, employs a different training objective and has demonstrated superior performance on several \ac{nlp} benchmarks. RoBERTa models are versatile, capable of handling both span and document classification tasks. In our Dutch clinical case study, we used MedRoBERTa.nl developed by Verkijk and Vossen \cite{Verkijk2022}, a RoBERTa model with $125$ million parameters trained from scratch on Dutch clinical notes from the Amsterdam University Medical Center.

\subsubsection{Recurrent networks}

The \ac{lstm}, \ac{gru}, and \ac{qrnn} are types of \acp{rnn} suitable for both span and document classification tasks, the latter simply being a special case of span classification. We used the \ac{bigru}, which processes sequences both from left-to-right and right-to-left. This bidirectional approach helps maintain context over longer token sequences. Known downsides of \acp{rnn} are the lack of parallelisation and the sensitivity to hyperparameter tuning, both of which result in a significant amount of computational resources required for training.

\subsubsection{Convolutional Neural Networks}

\Acp{cnn} are another powerful type of neural network, often used for span and document classification tasks. In our study, we utilised a bidirectional variant of \ac{cnn}, which processes text sequences in both forward and backward directions. This bidirectional approach helps capture context from both ends of the sequence, similar to bidirectional \acp{gru}. \Acp{cnn} excel at capturing local patterns in data, making them well-suited for text, where n-grams or small phrases can be crucial for understanding context. Unlike \acp{rnn}, \acp{cnn} can process data in parallel, significantly speeding up the training process.
\\ \\
The primary advantage of \acp{cnn} is their ability to efficiently capture spatial hierarchies in data through convolutional and pooling layers. However, they may struggle with maintaining long-range dependencies compared to \acp{rnn} like \ac{bigru}. Despite this, bidirectional \acp{cnn} are computationally efficient and less sensitive to hyperparameter tuning, making them a practical choice for many text classification tasks.

\subsection{Performance evaluation}

Span classification involves two distinct tasks, identifying spans and subsequently classifying them. Therefore, our performance evaluation included two aspects. We assessed span identification performance using a token-based coverage expressed using the Jaccard index. For span classification, we evaluated assuming the correct spans are identified. Additionally, we measured end-to-end performance, which combines both span identification and span classification. For both span and document classification, we reported weighted and macro precision, recall, and F1-score.

Finally, in clinical practice it is important to consider the number of false labels, i.e. the number of spans that are falsely labelled with \textit{any} class value (other than "\textit{no label}" or "\textit{normal}"). We present the rate of false labelling relative to the total number of identified spans for our span classification task.

\section{Results}

This section provides the performance scores on the span and document-level label extraction tasks.

\subsection{Span classification}

Table~\ref{tab:span_class_semantic_performance_pipeline} shows that for most characteristics, SpanCategorizer achieved the highest weighted and macro F1-scores for the span classification task. However, ALL$_{rule}$ performed particularly well in classifying valvular disorders. This high performance may be attributed to these disorders being often described with very short, distinct phrases (Table~\ref{tab:span_characteristics}). Conversely, the remaining characteristics are typically described using longer, less distinctive spans, where SpanCategorizer demonstrated a better performance. MedCAT demonstrated a lower precision and recall in the span classification task. These results may be due to an imperfect span suggestion. This hypothesis is supported by Table~\ref{tab:span_class_semantic_performance_matching_spans} and~\ref{tab:span_class_coverage_performance}, which illustrate a high performance in span classification when the exact spans containing a label are suggested, but a low Jaccard-index when comparing MedCAT's end-to-end predicted spans containing a label with the ground truth. In addition, Table~\ref{tab:span_class_false_positive} details that MedCAT has a high percentage of false positive span labels, leading to a reduced precision. This indicates that using MedCAT combined with a greedy span suggester could improve results even further.

\begin{table}[htbp]
\setlength\tabcolsep{4pt}
\footnotesize
\caption{Semantic end-to-end performance of span classification methods}
\label{tab:span_class_semantic_performance_pipeline}
    \begin{tabular*}{\textwidth}{p{2.5cm}|p{0.75cm}p{0.75cm}p{1cm}|p{0.75cm}p{0.75cm}p{1cm}|p{0.75cm}p{0.75cm}p{1cm}}
        &  \multicolumn{3}{c@{}|}{\textbf{SpanCategorizer}}    & \multicolumn{3}{c@{}|}{\textbf{MetaCAT}} & \multicolumn{3}{c@{}}{\textbf{ALL$_{rule}$}} \\
\textbf{Characteristic} & F1 & recall & precision & F1 & recall  & precision & F1 & recall & precision \\
\hline
\raggedright Aortic regurgitation & 0.90 (0.67) & 0.85 (0.62) & \textbf{0.94} (0.73) & 0.49 (0.46) & 0.54 (0.50) & 0.50 (0.46) & \textbf{0.92} (\textbf{0.89}) & \textbf{0.90} (\textbf{0.87}) & \textbf{0.94} (\textbf{0.92}) \\
\raggedright Aortic stenosis & 0.82 (0.74) & 0.79 (0.67) & \textbf{0.86} (\textbf{0.85}) & 0.45 (0.38) & 0.46 (0.51) & 0.43 (0.40) & \textbf{0.83} (\textbf{0.75}) & \textbf{0.84} (\textbf{0.75}) & 0.83 (0.77) \\
\raggedright Diastolic dysfunction & \textbf{0.87} (\textbf{0.83}) & \textbf{0.85} (\textbf{0.81}) & \textbf{0.90} (\textbf{0.86}) & 0.55 (0.66) & 0.69 (0.66) & 0.60 (0.65) & 0.60 (0.60) & 0.60 (0.59) & 0.61 (0.61) \\
\raggedright Left ventricular dilatation & \textbf{0.84} (\textbf{0.89}) & \textbf{0.82} (0.85) & \textbf{0.85} (\textbf{0.93}) & 0.57 (0.65) & 0.32 (0.46) & 0.40 (0.53) & 0.75 (0.85) & 0.81 (\textbf{0.86}) & 0.70 (0.86) \\
\raggedright Left ventricular systolic dysfunction & \textbf{0.77} (\textbf{0.42}) & \textbf{0.75} (0.41) & \textbf{0.79} (\textbf{0.43}) & 0.33 (0.24) & 0.69 (\textbf{0.49}) & 0.44 (0.32) & 0.21 (0.22) & 0.16 (0.19) & 0.33 (0.31) \\
\raggedright Mitral regurgitation & \textbf{0.93} (0.71) & 0.9 (0.69) & \textbf{0.97} (0.72) & 0.63 (0.76) & 0.59 (0.6) & 0.61 (0.66) & 0.92 (\textbf{0.91}) & \textbf{0.91} (\textbf{0.89}) & 0.93 (\textbf{0.92}) \\
\raggedright Pericardial effusion & \textbf{0.79} (0.28) & \textbf{0.70} (0.25) & \textbf{0.89} (\textbf{0.32}) & 0.66 (\textbf{0.35}) & 0.60 (\textbf{0.26}) & 0.62 (0.29) & 0.74 (0.21) & 0.62 (0.19) & 0.93 (0.26) \\
\raggedright Right ventricular dilatation & \textbf{0.90} (0.72) & \textbf{0.88} (0.71) & \textbf{0.93} (0.74) & 0.26 (0.44) & 0.23 (0.33) & 0.25 (0.37) & 0.77 (\textbf{0.85}) & 0.80 (\textbf{0.83}) & 0.75 (\textbf{0.88}) \\
\raggedright Right ventricular systolic dysfunction & \textbf{0.89} (\textbf{0.64}) & \textbf{0.88} (\textbf{0.66}) & \textbf{0.9} (0.62) & 0.61 (0.6) & 0.68 (0.51) & 0.64 (0.54) & 0.53 (0.48) & 0.37 (0.33) & 0.96 (\textbf{0.95}) \\
\raggedright Tricuspid regurgitation & 0.90 (\textbf{0.83}) & 0.88 (0.81) & \textbf{0.93} (\textbf{0.85}) & 0.38 (0.40) & 0.51 (0.58) & 0.41 (0.44) & \textbf{0.92} (\textbf{0.83}) & \textbf{0.93} (\textbf{0.91}) & 0.91 (0.82) \\
\raggedright Wall motion abnormalities & \textbf{0.60} (\textbf{0.60}) & \textbf{0.61} (\textbf{0.63}) & \textbf{0.59} (\textbf{0.59}) & 0.24 (0.24) & 0.51 (0.52) & 0.32 (0.32) & 0.16 (0.24) & 0.18 (0.25) & 0.15 (0.23) \\
    \end{tabular*}
\footnotetext{Weighted and macro (in brackets) scores. The highest performance for each characteristic is denoted in bold.}
\end{table}

\begin{table}[htbp]
\setlength\tabcolsep{4pt}
\footnotesize
\caption{Semantic performance of span classification methods, assuming matching spans}
\label{tab:span_class_semantic_performance_matching_spans}
\centering
    \begin{tabular*}{\textwidth}{p{2.5cm}|p{0.75cm}p{0.75cm}p{1cm}|p{0.75cm}p{0.75cm}p{1cm}|p{0.75cm}p{0.75cm}p{1cm}}
                             &  \multicolumn{3}{c|}{\textbf{SpanCategorizer}}    & \multicolumn{3}{c}
                             {\textbf{MetaCAT}} & \multicolumn{3}{c}{\textbf{ALL$_{rule}$}}\\
\textbf{Characteristic}               & F1 & recall & precision & F1 & recall  & precision & F1 & recall & precision  \\
         \hline
  \raggedright Aortic regurgitation                & 0.91 (0.54) & 0.85 (0.50) &  0.97 (0.58)             & \textbf{0.98} (0.92) & 0.98 (0.89) & 0.98 (0.94)                                                & 0.91 (\textbf{0.89}) & 0.86 (0.87) & 0.96 (\textbf{0.92}) \\
  \raggedright Aortic stenosis                     & 0.88 (0.63) & 0.79 (0.53) & \textbf{1.00} (0.78)     & \textbf{0.97} (\textbf{0.84}) & \textbf{0.98} (\textbf{0.79}) & 0.97 (\textbf{0.91})           & 0.84 (0.75) & 0.83 (0.75) & 0.85 (0.77) \\
  \raggedright Diastolic dysfunction               & 0.91 (0.70) & 0.85 (0.64) & \textbf{0.98} (0.77)     & \textbf{0.98} (\textbf{0.94}) & \textbf{0.98} (\textbf{0.95}) & 0.98 (0.93)                    & 0.60 (0.60) & 0.58 (0.59) & 0.62 (0.61) \\
  \raggedright Left ventricular dilatation         & 0.90 (0.76) & 0.82 (0.71) & \textbf{1.00} (0.83)     & \textbf{0.97} (\textbf{0.89}) & \textbf{0.97} (\textbf{0.9})  & 0.97 (\textbf{0.88})           & 0.82 (0.85) & 0.79 (0.86) & 0.85 (0.86) \\
  \raggedright Left ventricular systolic dysfunction  & 0.84 (0.41) & 0.75 (0.36) & \textbf{0.97} (0.49)     & \textbf{0.95} (\textbf{0.64}) & \textbf{0.95} (\textbf{0.65}) & 0.95 (\textbf{0.63})           & 0.20 (0.22) & 0.14 (0.18) & 0.37 (0.31) \\
  \raggedright Mitral regurgitation                & 0.93 (0.57) & 0.90 (0.55) & 0.96 (0.59)              & \textbf{0.98} (\textbf{0.94}) & \textbf{0.98} (\textbf{0.93}) & \textbf{0.98} (\textbf{0.95})  & 0.90 (0.91) & 0.86 (0.89) & 0.94 (0.92) \\
  \raggedright Pericardial effusion                & 0.76 (0.24) & 0.70 (0.21) & 0.85 (0.30)              & \textbf{0.97} (\textbf{0.56}) & \textbf{0.97} (\textbf{0.53}) & \textbf{0.97} (\textbf{0.66})  & 0.74 (0.21) & 0.62 (0.19) & 0.93 (0.26) \\
  \raggedright Right ventricular dilatation        & 0.93 (0.62) & 0.88 (0.59) & \textbf{0.98} (0.65)     & \textbf{0.98} (\textbf{0.92}) & \textbf{0.98} (\textbf{0.94}) & 0.98 (\textbf{0.91})           & 0.78 (0.85) & 0.76 (0.83) & 0.80 (0.88) \\
  \raggedright Right ventricular systolic dysfunction & 0.91 (0.54) & 0.88 (0.53) & 0.94 (0.55)              & \textbf{0.97} (\textbf{0.9})  & \textbf{0.97} (\textbf{0.85}) & \textbf{0.97} (\textbf{0.95})  & 0.52 (0.48) & 0.36 (0.33) & 0.97 (\textbf{0.95}) \\
  \raggedright Tricuspid regurgitation             & 0.92 (0.68) & 0.88 (0.65) & \textbf{0.97} (0.74)     & \textbf{0.98} (\textbf{0.9})  & \textbf{0.98} (\textbf{0.89}) & \textbf{0.98} (\textbf{0.91})  & 0.90 (\textbf{0.83}) & 0.89 (\textbf{0.84}) & 0.91 (\textbf{0.82}) \\
  \raggedright Wall motion abnormalities           & 0.75 (0.51) & 0.61 (0.42) & \textbf{0.99} (0.66)     & \textbf{0.98} (\textbf{0.96}) & \textbf{0.98} (\textbf{0.96}) & 0.98 (\textbf{0.96})           & 0.16 (0.24) & 0.18 (0.25) & 0.15 (0.23) \\
    \end{tabular*}
\footnotetext{Weighted and macro (in brackets) scores. The highest performance for each characteristic is denoted in bold.}
\end{table}

\begin{table}[htbp]    
\caption{Jaccard-index of span classification methods}
\label{tab:span_class_coverage_performance}
\centering
\begin{tabular}{c|c|c|c}               
\textbf{Characteristic}                       &  \textbf{SpanCategorizer}    & \textbf{MetaCAT} & \textbf{ALL$_{rule}$} \\
         \hline
  Aortic regurgitation               & 0.96             & 0.56 & \textbf{0.99} \\
  Aortic stenosis                    & \textbf{0.98}    & 0.47 & 0.96 \\
  Diastolic dysfunction              & \textbf{0.98}    & 0.78 & 0.85 \\
  Left ventricular dilatation        & \textbf{0.96}    & 0.47 & \textbf{0.96} \\
  Left ventricular systolic dysfunction & \textbf{0.95} & 0.74 & 0.84 \\
  Mitral regurgitation               & 0.99             & 0.64 & 0.99 \\
  Pericardial effusion               & \textbf{0.96}    & 0.76 & \textbf{0.96} \\
  Right ventricular dilatation       & \textbf{0.99}    & 0.32 & 0.93 \\
  Right ventricular systolic dysfunction& \textbf{0.99} & 0.75 & \textbf{0.99} \\
  Tricuspid regurgitation            & \textbf{0.99}    & 0.57 & \textbf{0.99} \\
  Wall motion abnormalities          & \textbf{0.88}    & 0.55 & 0.74 \\
    \end{tabular}
\footnotetext{The highest performance for each characteristic is denoted in bold.}
\end{table}

\begin{table}[htbp]
\caption{Fraction of false positive span labels}
\label{tab:span_class_false_positive}
\centering
\begin{tabular}{c|c|c|c}              
\textbf{Characteristic}                       &  \textbf{SpanCategorizer}    & \textbf{MetaCAT} & \textbf{ALL$_{rule}$} \\
         \hline
  Aortic regurgitation               & $<$0.01  & 0.13 & $<$0.01 \\
  Aortic stenosis                    & 0.01     & 0.12 & $<$0.01 \\
  Diastolic dysfunction              & 0.03     & 0.10 & 0.04 \\
  Left ventricular dilatation        & 0.01     & 0.05 & 0.01 \\
  Left ventricular systolic dysfunction & 0.01     & 0.76 & 0.02 \\
  Mitral regurgitation               & 0.01     & 0.12 & 0.01 \\
  Pericardial effusion               & $<$0.01  & 0.04 & $<$0.01 \\
  Right ventricular dilatation       & 0.02     & 0.11 & 0.01 \\
  Right ventricular systolic dysfunction& 0.03     & 0.08 & $<$0.01 \\
  Tricuspid regurgitation            & $<$0.01  & 0.13 & 0.01 \\
  Wall motion abnormalities          & 0.04     & 0.22 & 0.08 \\
    \end{tabular}
\end{table}

\subsection{Document classification}

Results for the document classification task are presented in Table~\ref{tab:doc_class_performance} and~\ref{tab:doc_class_performance_heuristics}. From these tables, MedRoBERTa.nl outperforms all other models on weighted and macro F1-score, precision, and recall. Indirect document classification using span classifiers resulted in a suboptimal performance, highlighting the added value of direct document classification models. BOW, our second baseline approach, performed quite well considering that we did not perform feature processing except \ac{tf-idf} and lemmatisation. An explanation might be that, because we are dealing with short staccato notes, containing little elaborations, and primarily containing statements of facts. Another reason may be that the number of negations is limited in echocardiogram reports. We also applied a document averaging of clinical word embeddings, but this was not favorable with respect to \ac{bow} with \ac{tf-idf}.
\\ \\
For MedRoBERTa, we applied a de-abbreviation step to investigate whether the presence of several abbreviations, combined with the relative brevity of the notes, would undermine the model's performance. MedRoBERTa is competitive with methods like \acp{bilstm}, especially in the case of larger contexts. However, we did not observe an improvement over the original texts. This could be due to the already high performance without de-abbreviation. For both \ac{bigru} and \ac{cnn} models, the use of de-abbreviations also did not impact the performance favorably. 
\\ \\
Additionally, we experimented with using pre-trained word vectors concatenated with the original trainable embedding layer for the \ac{cnn} and \ac{bigru} models. We did not see a significant improvement in performance, but the added embeddings did incur increased computational cost. The benefit of such pre-trained embeddings might be more noticeable with smaller training sizes, adding contextual information that the model might not learn from a small dataset alone. We also experimented with stacked dilated \acp{cnn} and TextCNN, again with no noticeable performance improvement while incurring increased computational cost.
\\ \\
SetFit performed well considering that we used about $10\%$ of the samples resulting in about $12.000$ contrastive examples. The sentence embeddings based on the BioLord2023 model are notably worse than the sentence embeddings based on the more generic RobBERTv2 model (Additional file 3). This can be explained by the fact that the SBERT model for RobBERTv2 was trained on a broad semantic range or sentences whereas BioLORD used the LORD training that seeks to maximize difference between medical concept definitions and i.e. is more suitable for named-entity-recognition tasks.
\\ \\
Retraining all models on a reduced label set improves performance markedly (Table~\ref{tab:doc_class_performance_3labels}). Using a further reduced label set only including the presence or absence of a mention of an characteristic yielded near-perfect results. This approach can be particularly useful in practical applications where high precision is required, and resources for manual data labelling are limited.

\begin{landscape}
\begin{table}[htbp]
\setlength\tabcolsep{3pt}
\footnotesize
\caption{Semantic performance of document classification methods}
\label{tab:doc_class_performance}
\begin{tabular}{p{2.5cm}|p{0.75cm}p{0.75cm}p{1cm}|p{0.75cm}p{0.75cm}p{1cm}|p{0.75cm}p{0.75cm}p{1cm}|p{0.75cm}p{0.75cm}p{1cm}|p{0.75cm}p{0.75cm}p{1cm}}
                             &  \multicolumn{3}{c|}{\textbf{BOW}} & \multicolumn{3}{c|}{\textbf{SetFit (RobBERT)}} & \multicolumn{3}{c}{\textbf{MedRoBERTa.nl}} & \multicolumn{3}{c}{\textbf{biGRU}} & \multicolumn{3}{c}{\textbf{CNN}}\\
\textbf{Characteristic}               & F1 & recall & precision & F1 & recall & precision & F1 & recall & precision  & F1 & recall & precision & F1 & recall & precision  \\
         \hline
\raggedright Aortic regurgitation                  & 0.90 (0.74) & 0.90 (0.65) & 0.90 (0.84) & 0.93 (0.86) & 0.93 (0.88) & 0.93 (0.84) & \textbf{0.96} (\textbf{0.93}) & 0.95 (\textbf{0.90}) & \textbf{0.96} (\textbf{0.96}) & 0.94 (0.89) & 0.94 (0.88) & 0.93 (0.90) & 0.95 (0.89) & 0.95 (0.85)& 0.95 (0.95) \\
\raggedright Aortic stenosis                       & 0.95 (0.77) & 0.93 (0.72) & 0.93 (0.89) & 0.91 (0.82) & 0.91 (0.93) & 0.91 (0.75) & \textbf{0.96} (0.89) & \textbf{0.95} (0.91) & \textbf{0.96} (0.88) & 0.94 (0.88) & 0.94 (0.87) & 0.94 (0.89) & 0.94 (\textbf{0.91}) & 0.94 (\textbf{0.93}) & 0.94 (0.89)\\
  \raggedright Diastolic dysfunction               & 0.93 (0.82) & 0.93 (0.80) & 0.93 (0.84) & 0.95 (0.91) & 0.95 (\textbf{0.97}) & 0.95 (0.87) & \textbf{0.97} (\textbf{0.95})&\textbf{0.97} (0.96) & \textbf{0.98} (\textbf{0.94}) & 0.93 (0.84) & 0.93 (0.82) & 0.93 (0.86) & 0.94 (0.93) & 0.93 (0.93) & 0.94 (0.92)\\
  \raggedright Left ventricular dilatation         & 0.86 (0.56) & 0.86 (0.51) & 0.85 (0.63) & 0.95 (0.91) & 0.95 (\textbf{0.95}) & 0.95 (0.87) & \textbf{0.96} (\textbf{0.95})&\textbf{0.96} (\textbf{0.95}) & \textbf{0.97} (0.95) & 0.94 (0.90) & 0.93 (0.87) & 0.94 (\textbf{0.96}) & 0.94 (0.93) & 0.93 (0.93) & 0.94 (0.92)\\
  \raggedright Left ventricular systolic dysfunction  & 0.89 (0.82) & 0.89 (0.80) & 0.89 (0.84) & 0.95 (0.91) & 0.95 (0.92) & 0.95 (0.89) & \textbf{0.97} (\textbf{0.93}) & \textbf{0.96} (\textbf{0.92}) & \textbf{0.97} (\textbf{0.95}) & 0.93 (0.89) & 0.93 (0.89) & 0.93 (0.91) & 0.95 (0.92) & 0.95 (0.91) & 0.95 (0.92)\\
  \raggedright Mitral regurgitation                & 0.88 (0.68) & 0.88 (0.65) & 0.88 (0.74) & 0.94 (0.88) & 0.94 (0.93) & 0.94 (0.85) & \textbf{0.96} (0.92) & \textbf{0.96} (\textbf{0.94}) & \textbf{0.97} (0.90) & 0.92 (0.87) & 0.92 (0.85) & 0.93 (0.88) & 0.94 (0.92) & 0.94 (0.92) & 0.94 (0.93)\\
  \raggedright Pericardial effusion                & 0.95 (0.42) & 0.95 (0.40) & 0.94 (0.48) & 0.92 (0.51) & 0.92 (0.60) & 0.92 (0.49) & \textbf{0.98} (\textbf{0.81}) & \textbf{0.98} (\textbf{0.80}) & 0.97 (0.84) & 0.97 (0.75) & 0.96 (0.69) & \textbf{0.98} (\textbf{0.86}) & 0.97 (0.63) & 0.97 (0.60) & 0.97 (0.69)\\
  \raggedright Right ventricular dilatation        & 0.86 (0.54) & 0.87 (0.49) & 0.86 (0.68) & 0.92 (0.81) & 0.92 (0.92) & 0.92 (0.74) & \textbf{0.96} (\textbf{0.95}) & \textbf{0.96} (\textbf{0.96}) & \textbf{0.96} (\textbf{0.94}) & 0.93 (0.91) & 0.93 (0.90) & 0.94 (0.93) & 0.94 (0.89) & 0.94 (0.88) & 0.94( 0.89)\\
  \raggedright Right ventricular systolic dysfunction & 0.89 (0.66) & 0.89 (0.64) & 0.89 (0.75) & 0.94 (0.89) & 0.94 (0.93) & 0.94 (0.85) & \textbf{0.96} (\textbf{0.93}) & \textbf{0.96} (\textbf{0.94}) & \textbf{0.96} (\textbf{0.92}) & 0.92 (0.76) & 0.91 (0.77) & 0.92 (0.75) & 0.91 (0.78) & 0.90 (0.75) & 0.92 (0.84))\\
  \raggedright Tricuspid regurgitation             & 0.90 (0.64) & 0.90 (0.63) & 0.90 (0.79) & 0.92 (0.83) & 0.92 (0.86) & 0.92 (0.80) & 0.96 (0.92) & 0.96 (0.92) & 0.96 (0.92) & 0.95 (0.92) & 0.95 (0.91) & 0.95 (0.94) & 0.96 (0.95) & 0.96 (\textbf{0.95}) & 0.96 (0.94)\\
  \raggedright Wall motion abnormalities           & 0.95 (0.90) & 0.95 (0.87) & 0.95 (0.93) & 0.95 (0.92) & 0.95 (0.93) & 0.95 (0.91) & \textbf{0.97} (\textbf{0.95}) & \textbf{0.97} (\textbf{0.95}) & \textbf{0.97} (\textbf{0.96}) & 0.95 (0.92) & 0.95 (0.90) & 0.95 (0.93) & 0.96 (0.94) & 0.96 (0.92) & 0.96 (\textbf{0.96})\\
    \end{tabular}
\footnotetext{Weighted and macro (in brackets) scores. The highest performance for each characteristic is denoted in bold.}
\end{table}
\end{landscape}

\begin{table}[htbp]
\setlength\tabcolsep{4pt}
\footnotesize
\caption{Semantic performance of span $\rightarrow$ document classification heuristics}
\label{tab:doc_class_performance_heuristics}
\centering
    \begin{tabular}{p{2.5cm}|p{0.75cm}p{0.75cm}p{1cm}|p{0.75cm}p{0.75cm}p{1cm}|p{0.75cm}p{0.75cm}p{1cm}}
                             &  \multicolumn{3}{c|}{\textbf{SpanCategorizer}} & \multicolumn{3}{c|}{\textbf{MedCAT}} & \multicolumn{3}{c}{\textbf{ALL$_{rule}$}}\\
\textbf{Characteristic}               & F1 & recall & precision & F1 & recall & precision & F1 & recall & precision \\
         \hline
  \raggedright Aortic regurgitation                & \textbf{0.95} (0.74) & \textbf{0.95} (0.71) & \textbf{0.95} (0.77) & 0.6 (0.46) & 0.58 (0.56) & 0.63 (0.42) & \textbf{0.95} (\textbf{0.91}) & \textbf{0.95} (\textbf{0.88}) & \textbf{0.95} (\textbf{0.95})  \\
  \raggedright Aortic stenosis                     & 0.94 (0.83) & 0.94 (0.76) & 0.94 (0.93) & 0.64 (0.48) & 0.62 (0.66) & 0.67 (0.42) & \textbf{0.95} (\textbf{0.9}) & \textbf{0.95} (\textbf{0.86}) & \textbf{0.95} (\textbf{0.96})  \\
  \raggedright Diastolic dysfunction               & \textbf{0.94} (\textbf{0.87}) & \textbf{0.94} (\textbf{0.84}) & \textbf{0.94} (0.91) & 0.75 (0.67) & 0.74 (0.77) & 0.81 (0.62) & 0.93 (0.82) & 0.93 (0.76) & 0.93 (\textbf{0.93})  \\
  \raggedright Left ventricular dilatation         & 0.91 (0.59) & 0.91 (0.59) & 0.91 (0.6) & 0.65 (0.54) & 0.67 (0.58) & 0.69 (0.54) & \textbf{0.94} (\textbf{0.91}) & \textbf{0.94} (\textbf{0.89}) & \textbf{0.94} (\textbf{0.94})  \\
  \raggedright Left ventricular systolic dysfunction  & \textbf{0.92} (\textbf{0.88}) & \textbf{0.92} (\textbf{0.89}) & \textbf{0.92} (\textbf{0.89}) & 0.87 (0.79) & 0.86 (0.8) & 0.88 (0.81) & 0.33 (0.37) & 0.33 (0.41) & 0.33 (0.75)  \\
  \raggedright Mitral regurgitation                & \textbf{0.96} (\textbf{0.92}) & \textbf{0.96} (\textbf{0.9}) & \textbf{0.96} (\textbf{0.95}) & 0.67 (0.64) & 0.67 (0.65) & 0.68 (0.64) & 0.94 (\textbf{0.92}) & 0.94 (\textbf{0.9}) & 0.94 (0.94)  \\
  \raggedright Pericardial effusion                & 0.95 (0.32) & 0.95 (0.32) & 0.95 (0.32) & 0.87 (\textbf{0.48}) & 0.85 (\textbf{0.55}) & 0.9 (\textbf{0.6}) & \textbf{0.96} (0.37) & \textbf{0.96} (0.46) & \textbf{0.96} (0.36)  \\
  \raggedright Right ventricular dilatation        & \textbf{0.93} (0.72) & \textbf{0.93} (0.68) & \textbf{0.93} (0.78) & 0.64 (0.43) & 0.63 (0.49) & 0.67 (0.44) & 0.9 (\textbf{0.83}) & 0.9 (\textbf{0.8}) & 0.9 (\textbf{0.88})  \\
  \raggedright Right ventricular systolic dysfunction & \textbf{0.94} (\textbf{0.72}) & \textbf{0.94} (\textbf{0.75}) & \textbf{0.94} (0.7) & 0.78 (0.63) & 0.78 (0.67) & 0.8 (0.6) & 0.72 (0.55) & 0.72 (0.47) & 0.72 (\textbf{0.89})  \\
  \raggedright Tricuspid regurgitation             & \textbf{0.96} (0.92) & \textbf{0.96} (0.9) & \textbf{0.96} (0.96) & 0.6 (0.48) & 0.57 (0.69) & 0.68 (0.43) & \textbf{0.96} (\textbf{0.97}) & \textbf{0.96} (\textbf{0.96}) & \textbf{0.96} (\textbf{0.98})  \\
  \raggedright Wall motion abnormalities           & \textbf{0.95} (\textbf{0.92}) & \textbf{0.95} (0.9) & \textbf{0.95} (\textbf{0.96}) & 0.55 (0.45) & 0.52 (0.61) & 0.77 (0.48) & \textbf{0.95} (\textbf{0.92}) & \textbf{0.95} (\textbf{0.93}) & \textbf{0.95} (0.91)  \\
    \end{tabular}
\footnotetext{Weighted and macro (in brackets) scores. The highest performance for each characteristic is denoted in bold.}
\end{table}

\begin{landscape}
\begin{table}[htbp]
\setlength\tabcolsep{4pt}
\footnotesize
\caption{Semantic performance of document classification methods for simplified label scheme (No label, Normal, and Present)}
\label{tab:doc_class_performance_3labels}
    \begin{tabular}{p{2.5cm}|p{0.75cm}p{0.75cm}p{1cm}|p{0.75cm}p{0.75cm}p{1cm}|p{0.75cm}p{0.75cm}p{1cm}|p{0.75cm}p{0.75cm}p{1cm}|p{0.75cm}p{0.75cm}p{1cm}}
                             &  \multicolumn{3}{c|}{\textbf{BOW}} & \multicolumn{3}{c|}{\textbf{SetFit (RobBERT)}} & \multicolumn{3}{c}{\textbf{MedRoBERTa.nl}} & \multicolumn{3}{c}{\textbf{biGRU}} & \multicolumn{3}{c}{\textbf{CNN}}\\
\textbf{Characteristic}               & F1 & recall & precision & F1 & recall & precision & F1 & recall & precision  & F1 & recall & precision & F1 & recall & precision \\
         \hline
  \raggedright Aortic regurgitation                & 0.92 (0.89) & 0.92 (0.88) & 0.92 (0.89) & 0.94 (0.93) & 0.94 (0.94) & 0.94 (0.91)  & 0.97 (0.97) & 0.97 (0.97) & 0.97 (0.97) & 0.96 (0.95) & 0.96 (0.95) & 0.96 (0.96) & 0.96 (0.96) & 0.96 (0.95) & 0.96 (0.96) \\
  \raggedright Aortic stenosis                     & 0.94 (0.89) & 0.94 (0.88) & 0.94 (0.90) & 0.91 (0.86) & 0.91 (0.94) & 0.91 (0.82) & 0.95 (0.93) &0.95 (0.95) & 0.95 (0.93) & 0.95 (0.93) & 0.95 (0.94) & 0.95 (0.92) & 0.96 (0.94) & 0.96 (0.95) & 0.96 (0.95) \\
  \raggedright Diastolic dysfunction               & 0.94 (0.91) & 0.94 (0.90) & 0.94 (0.91) & 0.95 (0.92) & 0.95 (\textbf{0.97}) & 0.95 (0.89) &\textbf{0.97} (\textbf{0.96}) & \textbf{0.97} (\textbf{0.97}) & \textbf{0.97} (\textbf{0.96}) & 0.96 (0.94) & 0.96 (0.95) & 0.96 (0.92) & 0.96 (0.95) & 0.96 (0.95) & \textbf{0.97} (0.95)\\
  \raggedright Left ventricular dilatation         & 0.88 (0.82) & 0.88 (0.81) & 0.88 (0.84) & 0.95 (0.94) & 0.95 (0.96) & 0.95 (0.93) & \textbf{0.96} (0.94) & \textbf{0.96} (\textbf{0.95}) & \textbf{0.96} (0.94) & 0.95 (0.94) & 0.95 (0.94) & 0.95 (0.94) & \textbf{0.96} (\textbf{0.95}) & \textbf{0.96} (\textbf{0.95}) & \textbf{0.96} (\textbf{0.95})\\
  \raggedright Left ventricular systolic dysfunction  & 0.92 (0.90) & 0.92 (0.89) & 0.92 (0.90) & 0.96 (0.94) & 0.96 (\textbf{0.95}) & 0.96 (0.93) & \textbf{0.97} (\textbf{0.95}) & \textbf{0.97} (\textbf{0.95}) & \textbf{0.97} (\textbf{0.95}) & 0.96 (0.94) & 0.96 (0.94) & \textbf{0.97} (0.94) & 0.96 (0.94) & 0.96 (0.93) & 0.96 (0.94)\\
  \raggedright Mitral regurgitation                & 0.90 (0.88) & 0.90 (0.88) & 0.90 (0.89) & 0.94 (0.94) & 0.94 (0.95) & 0.94 (0.93) & \textbf{0.97} (\textbf{0.97}) & \textbf{0.97} (\textbf{0.97}) & \textbf{0.97} (0.96) & 0.95 (0.94) & 0.94 (0.95) & 0.95 (0.94) & 0.96 (0.95) & 0.96 (0.96) & 0.96 (0.95)\\
  \raggedright Pericardial effusion                & 0.96 (0.84) & 0.97 (0.82) & 0.96 (0.88) & 0.95 (0.85) & 0.95 (0.93) & 0.95 (0.79) & \textbf{0.99} (\textbf{0.95}) & \textbf{0.99} (\textbf{0.96}) & \textbf{0.99} (0.94) & 0.98 (0.94) & 0.98 (0.95) & 0.98 (0.94) & 0.98 (0.94) & 0.98 (0.94) & 0.98 (\textbf{0.95}) \\
  \raggedright Right ventricular dilatation        & 0.87 (0.79) & 0.88 (0.77) & 0.87 (0.81) & 0.91 (0.86) & 0.91 (0.91) & 0.91 (0.83) & \textbf{0.95} (\textbf{0.93}) & \textbf{0.95} (\textbf{0.95}) & \textbf{0.96} (0.92) & 0.94 (0.92) & 0.94 (0.92) & 0.94 (0.92) & \textbf{0.95} (0.92) & \textbf{0.95} (0.93) & 0.95 (0.92)\\
  \raggedright Right ventricular systolic dysfunction & 0.91 (0.86) & 0.90 (0.85) & 0.90 (0.87) & 0.93 (0.91) & 0.93 (0.94) & 0.93 (0.90) & \textbf{0.97} (\textbf{0.95}) & \textbf{0.97} (\textbf{0.94}) & \textbf{0.97} (\textbf{0.96}) & 0.94 (0.92) & 0.94 (0.91) & 0.94 (0.93) & 0.94 (0.92) & 0.94 (0.92) & 0.95 (0.93) \\
  \raggedright Tricuspid regurgitation             & 0.93 (0.90) & 0.93 (0.90) & 0.93 (0.89) & 0.93 (0.91) & 0.93 (0.93) & 0.93 (0.89) & \textbf{0.97} (\textbf{0.97}) & \textbf{0.97} (0.96) & \textbf{0.97} (\textbf{0.97}) & 0.96 (0.95) & 0.96 (0.95) & 0.96 (0.95) & \textbf{0.97} (0.96) & \textbf{0.97} (0.96) & \textbf{0.97} (0.95)\\
  \raggedright Wall motion abnormalities           & 0.94 (0.90) & 0.94 (0.85) & 0.94 (0.92) & 0.95 (0.92) & 0.95 (0.93) & 0.95 (0.91) & \textbf{0.97} (\textbf{0.95}) & \textbf{0.97} (\textbf{0.94}) & \textbf{0.97} (\textbf{0.96}) & 0.96 (0.93) & 0.96 (0.90) & 0.96 (\textbf{0.96}) & \textbf{0.97} (0.94) & \textbf{0.97} (0.93) & \textbf{0.97} (\textbf{0.96})\\
\end{tabular}
\footnotetext{Weighted and macro (in brackets) scores. The highest performance for each characteristic is denoted in bold.}
\end{table}
\end{landscape}

\section{Discussion}
This study aimed to explore and compare various NLP methods for extracting clinical labels from unstructured Dutch echocardiogram reports. We developed and evaluated several approaches for both span- and document-level label extraction on an internal test set, demonstrating high performance in identifying eleven commonly described cardiac characteristics, including left and right ventricular systolic dysfunction, left and right ventricular dilatation, diastolic dysfunction, aortic stenosis, aortic regurgitation, mitral regurgitation, tricuspid regurgitation, pericardial effusion, and wall motion abnormalities. The main findings indicate that SpanCategorizer consistently outperformed other models in span-level classification tasks, achieving weighted F1-scores ranging from $0.60$ to $0.93$ across these characteristics, while MedRoBERTa.nl excelled in document-level classification with a weighted F1-score exceeding $0.96$ for all characteristics. 
\\ \\
In this study, we observed a variation in results of different span classification approaches. The baseline approach, using regular expressions, achieved a high performance for some characteristics but performed poorly for others. These outcomes are likely linked to span length, frequency, and distinctiveness \cite{papay2020}. Our most poorly performing characteristics - left ventricular systolic dysfunction, pericardial effusion, and wall motion abnormalities - have larger span lengths, and lower span frequencies. Macro performance is particularly impacted by the ‘severe’ classes, which have a low span frequency and high span length, which have both been previously linked to worse performance \cite{papay2020}. 
\\ \\
The MedCAT approach has a very high overall precision but lacked recall due to imperfect span suggestions. Therefore, for medical applications, it may be more effective to use a greedy span-classifier as the primary span suggestion method, with a \ac{ner}+L extraction serving as an augmentation tool to extract additional features. Alternatively, to make the MedCAT model more robust, we should consider using fuzzy matching with varying proximities, using tools like clinlp \cite{menger_2024}, instead of adding possible spans directly from the training phase of the labelling process in Prodigy. Another approach, given the results from the document classification task, could involve training a RoBERTa-based or \ac{cnn}/\ac{bigru} span classifier, using either a MedCAT or greedy span suggester. Additionally, a joint entity/relation extraction model could be constructed \cite{eberts_span-based_2020}. However, these approaches are outside the scope of the current paper and require significantly higher computational cost.
\\ \\
For document classification, the MedRoBERTa.nl model demonstrated the best overall performance. This aligns with previous findings, which highlight the additional value of BERT-based models in cases involving infrequently occurring spans \cite{papay2020}. We did not attempt to train a BERT-based model from scratch due to the limited number of available documents. Previous studies have shown that pre-training on a small corpus yields suboptimal results, whereas models with general domain pre-training, such as MedRoBERTa.nl, achieve highly competitive results without requiring domain-specific feature engineering \cite{garcia-pablos_sensitive_2020, madabushi_cost-sensitive_2020, limsopatham_effectively_2021, muizelaar_extracting_2024, rietberg_accurate_2023, tucker_transformers_2021}. The \ac{bigru} and \ac{cnn} models demonstrated a competitive performance, especially considering their significantly lower computational cost. Alternatives like TextCNN or hierarchical architectures such as Hierarchical Attention Networks might perform better with longer contexts, such as discharge summaries \cite{kim_convolutional_2014, yang_hierarchical_2016}. 
\\ \\
The \ac{bow} approach, while effective considering its simplicity, could have been extended with more sophisticated weighting mechanisms, such as incorporating negation estimation, part-of-speech tagging, and dependency parsing. These additions could have improved the contextual understanding of the text, potentially leading to better document classification. However, such extensions would require significantly more complex feature engineering and computational resources, which were beyond the scope of this study.
\\ \\
Regarding the SetFit method, three remarks can be made. First, training a new sentence transformer from scratch based on the MedRoBERTa.nl model might yield better results than using the arithmetic mean. Second, the BioLORD model is constrastively trained to discriminate between medical span-level concepts, rather than explicitly between semantic differences. Third, we achieved performance close to the best-performing method using only $10\%$ of the data. Therefore, this approach may be most suitable given the resources required for manual data labelling.
\\ \\
The class distribution in our dataset reflects real-world practice, with over $75$\% of documents lacking a label for at least one characteristic, and a small percentage containing moderate or severe labels. While this distribution poses challenges for model performance, particularly in terms of macro scores, it also highlights the need for models to perform well under realistic clinical conditions. Expanding the dataset was not feasible due to the extensive manual labeling process, which already took several months. An alternative approach to enhance model performance could involve utilizing English BERT-based models on translated texts, as suggested by Muizelaar et al. \cite{muizelaar_extracting_2024}.
\\ \\
We employed a single train/test split for our experiments, which, while practical, could introduce certain limitations. One potential concern is the risk of overfitting to the specific data in the training set, particularly when using handcrafted features like regular expressions. This might result in models that perform well on the test set but may not generalize as effectively to new, unseen data. Ideally, a cross-validation approach would provide a more comprehensive evaluation by averaging performance across multiple splits, thereby reducing the variance and offering a more robust assessment of model performance. However, given the infeasibility of developing regular expressions for each fold, our approach represents a pragmatic balance between practical constraints and methodological rigor. The use of a single split also means that our performance estimates may be somewhat optimistic, as they are tied to the specific characteristics of the selected test set. This is particularly relevant for our span classification tasks, where the performance varied significantly across different span types. In future work, incorporating cross-validation or a more extensive test set could help mitigate these limitations, providing a clearer picture of how well these models might perform in broader clinical applications.
\\ \\
Our findings suggest distinct use cases for span and document classification within clinical practice. Span classification, while adding a layer of explainability by highlighting specific spans that contribute to a particular label, exhibit too much variability in performance to be reliably used in clinical settings. This inconsistency, especially across different characteristics, limits its utility for direct clinical application at this stage. In contrast, document classification demonstrated significantly better and more consistent performance, making it a more viable option for integration into clinical workflows. This approach could be effectively used for tasks such as constructing patient cohorts for research or automating parts of the diagnostic pipeline. Additionally, we observed that reducing the number of labels significantly improved the performance of document classification models. This reduced label model might be employed to flag cases that require more detailed review, either by activating a clinician's attention or by supporting active labeling in research settings, such as using Prodigy. This approach not only enhances model accuracy but also provides a practical pathway for implementing \ac{nlp} tools in clinical environments where efficiency and precision are essential.

\section*{Conclusions}

In this study, we developed several \ac{nlp} methods for span- and document-level label extraction from Dutch unstructured echocardiogram reports and evaluated the performance of these methods on an internal test set. We demonstrate high performance in identifying eleven cardiac characteristics. Performance for span classification ranges between a weighted F1-score of $0.60$ and $0.93$ for all characteristics using the SpanCategorizer model, while for document classification we achieve a weighted F1-score of $>$$0.96$ for all characteristics using the MedRoBERTa.nl model. 
\\ \\
For future research, we suggest to use the SpanCategorizer and MedRoBERTa.nl model for span- and document-level diagnosis extraction from Dutch unstructured echocardiogram reports, respectively. In case of a limited amount of data, SetFit may be a suitable alternative for document classification. SpanCategorizer and MedRoBERTa.nl have been made publicly available through HuggingFace. Future work may include validation in external institutions or the extension to other cardiac characteristics.

\section*{List of abbreviations}
\printacronyms[heading=none]

\section*{Declarations}

\subsection*{Ethics approval and consent to participate}
The \ac{umcu} quality assurance research officer confirmed under project number 22U-0292 that this study does not fall under the scope of the Dutch Medical Research Involving Human Subjects Act (WMO) and therefore does not require approval from an accredited medical ethics committee. The study was performed compliant with local legislation and regulations. The need for patient consent was waived, as subjects are not actively involved in this study. All patient data used was used after pseudonomisation.

\subsection*{Consent for publication}
Not applicable.

\subsection*{Availability of data and materials}
The datasets generated and/or analysed during the current study are not publicly available due to potential privacy-sensitive information, but are available from the corresponding author upon reasonable request and local institutional approval. Research code is publicly available on GitHub, via \href{https://github.com/umcu/EchoLabeler}{https://github.com/umcu/EchoLabeler}.

\subsection*{Competing interests}
The authors declare that they have no competing interests.

\subsection*{Funding}
The work received funding from the European Union's Horizon Europe research
and innovation programme under Grant Agreement No. 101057849 (DataTools4Heart project). 
The collaboration project is co-funded by PPP Allowance awarded by Health$\sim$Holland,
Top Sector Life Sciences \& Health, to stimulate public-private partnerships. 
This publication is part of the project MyDigiTwin with project number $628.011.213$ of the research 
programme COMMIT2DATA Big Data \& Health which is partly financed by the Dutch Research Council (NWO).

\subsection*{Authors' contributions}
B.A., M.V., B.E., and R.E. were responsible for conceptualisation. Data curation and manual labelling was performed by B.A. and M.V.. Methodology was developed by B.E., B.A. and M.V.. P.H. and A.T. were responsible for project supervision and administration. B.E., B.A., and M.V. performed experiments and validated the results. A.T. and D.O. gave clinical input on endpoint definitions. The original draft of this manuscript was written by B.A. and B.E.. All authors read and approved the final manuscript.

\subsection*{Acknowledgements}
The authors thank Celina Berkhoff for her contribution to the manual labelling process. Figures were created with BioRender.com.

\bibliography{bibliography} 

\appendix

\section{Additional file 1: Model parameters}
\subsection{biGRU \& CNN}

\begin{itemize}
\item Code libraries:
    \subitem Models: 
        \subsubitem Keras, \textit{version 2.15.0}
        \subsubitem TensorFlow, \textit{version 2.15.0}
    \subitem Support:
        \subsubitem Numpy, \textit{version 1.26.4}
        \subsubitem scitkit-learn, \textit{version 1.5.0}
        \subsubitem Pandas, \textit{version 2.2.2}
\item Parameters for biGRU
    \subitem vocabulary size: $5000$
    \subitem number of epochs: $20$
    \subitem model depth: $96$
    \subitem hidden layers FC: $10$
    \subitem maximum sequence length: $200$
    \subitem learning rate: $0.005$
    \subitem batch size: $128$
    \subitem embedding dimensions: $300$
    \subitem dropout ratio: $0.2$
    \subitem \textit{total learnable parameters}: $1,731,211$
\item Parameters for CNN
    \subitem vocabulary size: $5000$
    \subitem number of epochs: $20$
    \subitem model depth: $96$
    \subitem hidden layers FC: $10$
    \subitem maximum sequence length: $200$
    \subitem learning rate: $0.0025$
    \subitem batch size: $128$
    \subitem embedding dimensions: $300$
    \subitem dropout ratio: $0.2$
    \subitem \textit{total learnable parameters}: $1,645,099$
\end{itemize}

\subsection{MedRoBERTa.nl}

\begin{itemize}
\item Code libraries:
    \subitem Model: Transformers, \textit{version 4.40.0}
    \subitem Support: 
        \subsubitem Numpy, \textit{version 1.26.4}
        \subsubitem PyTorch, \textit{3.10.5}
\item Parameters for SetFit:
    \subitem loss: categorical cross entropy,
    \subitem Number of epochs: $5$,
    \subitem learning rate = $5e-5$,
    \subitem weight decay = $0.01$,
    \subitem batch size: $16$,
    \subitem seed: $42$,  
    \subitem \textit{total learnable parameters}: $125,980,419$
\item Parameters for SVC:
    \subitem \textit{defaults}
\end{itemize}

\subsection{SetFit}
\begin{itemize}
\item Code libraries:
    \subitem Embeddings: SentenceTransformers, \textit{version 3.0.0}
    \subitem Head: sklearn.svm.svc, \textit{version 1.5.0}
    \subitem Combination: SetFit, \textit{version 1.0.3}
    \subitem Support: 
        \subsubitem Transformers, \textit{version 4.40.0}
        \subsubitem Numpy, \textit{version 1.26.4}
        \subsubitem PyTorch, \textit{3.10.5}
\item Parameters for SetFit:
    \subitem loss: cosine similarity loss,
    \subitem Number of epochs: $3$,
    \subitem Number of samples: $500$,
    \subitem Number of iterations: $6$,
    \subitem body learning rate = $[2e-5, 1e-5]$,
    \subitem head learning rate= $1e-3$,
    \subitem batch size: $32$,
    \subitem seed: $42$,  
    \subitem \textit{total learnable parameters}: $118,892,544$
\item Parameters for SVC:
    \subitem \textit{defaults}
\end{itemize}

\subsection{BOW}
\begin{itemize}
\item Code libraries:
    \subitem LDA, TF-IDF: scikit-learn, \textit{version 1.5.0}
    \subitem XGBoost: xgboost, \textit{version 2.0.3}
    \subitem ETM: custom, see our git repository\footnote{\href{https://github.com/umcu/Echolabeller/blob/main/src/echo_utils.py}{$https://github.com/umcu/Echolabeller/blob/main/src/echo\_utils.py$}}
\item Parameters for LDA:
    \subitem number of topics: $20$
\item Parameters for TF-IDF:
    \subitem vocabulary size: $5000$
\item Parameters for XGBoost:
    \subitem seed: $42$
    \subitem number of estimators: $150$
    \subitem maximum depth: $5$
    \subitem learning rate: $1e-1$
    \subitem \textit{defaults otherwise}
\end{itemize}

\subsection{SpanCategorizer}
\begin{itemize}
\item Code libraries and model components:
    \subitem spacy, \textit{version 3.7.4}
    \subitem Embeddings: MultiHashEmbed.v2, 
        \subsubitem Attrs: ["NORM","PREFIX","SUFFIX","SHAPE"]
        \subsubitem Rows: [$5000$,$1000$,$2500$,$2500$]
    \subitem Encoding: spacy.MaxoutWindowEncoder.v2, 
        \subsubitem Width: $96$ 
        \subsubitem Depth: $4$ 
        \subsubitem Window size: $1$
        \subsubitem Maxout pieces: $3$
    \subitem Tok2vec: spacy.Tok2Vec.v2
\item Parameters for training:
    \subitem dropout: $0.1$
    \subitem max steps: $20,000$
    \subitem patience: $1,600$
    \subitem optimizer: Adam.v1
    \subitem learning rate: $0.001$
    \subitem score weights: $100$\% F1-score
\end{itemize}    

\subsection{MetaCAT}
\begin{itemize}
\item Code libraries and model components:
    \subitem MedCAT, \textit{version $1.10.2$}
    \subitem Embeddings: Skipgram on medical corpus, 
        \subsubitem dimension: $300$
        \subsubitem vocab size: $50,000$
\item Parameters for training:
    \subitem Model depth: $3$
    \subitem hidden layers FC: $256$
    \subitem Number of epochs: $25$
    \subitem Context window left of concept: $15$
    \subitem Context window right of concept: $10$
    \subitem learning rate: $1e-3$
    \subitem batch size: $128$
    \subitem dropout ratio: $0.25$
    \subitem \textit{total learnable parameters}: $16,233,975$
\end{itemize} 

\section{Additional file 2: Labelling instructions} \label{sec:label_instructions}
\subsection{General instructions}
Synonyms for categories:
\begin{itemize}
    \item Normal: trivial, trace
    \item Mildly abnormal: small, light, limited, grade I
    \item Moderately abnormal: grade II 
    \item Severely abnormal: important, massive, critical, large, grade III$+$
\end{itemize}
A correct span consists of the shortest amount of tokens where both characteristic (i.e., left ventricle, valve) and function (i.e., normal, mildly reduced) are included.\\
In case two severities for one characteristic are mentioned (i.e., moderate-severe), one span for each severity is created.
In general, the interpretation of the cardiologist (i.e., severe aortic stenosis) is annotated. Measurements (i.e., Vmax $5.0$ m/s) are not annotated.

\subsection{Left ventricular systolic dysfunction}
Since left ventricular ejection fraction (LVEF) is often only reported using percentages, we annotate the measurements in this case.
\begin{itemize}
    \item Normal: LVEF $>$$50$\%
    \item Mildly abnormal: LVEF $40$-$50$\%
    \item Moderately abnormal: LVEF $30$-$40$\%
    \item Severely abnormal: LVEF $\leq$$30$\%
\end{itemize}

\subsection{Left and right ventricular dilatation}
In cases of \textit{'normal dimensions'}, we label LV dilatation and RV dilatation as normal.

\subsection{Aortic stenosis}
In cases of \textit{'normal opening of the aortic valve'}, we label aortic stenosis as normal. Aortic sclerosis without aortic stenosis is marked as normal.

\subsection{Diastolic dysfunction}
\begin{itemize}
    \item Normal: normal diastolic function (for the patients' age)
    \item Mildly abnormal: grade I diastolic dysfunction, impaired ventricular relaxation
    \item Moderately abnormal: grade II diastolic dysfunction, pseudonormalisation
    \item Severely abnormal: grade III$+$ diastolic dysfunction, restrictive diastolic function
\end{itemize}

\subsection{Pericardial effusion}
The largest diameter of pericardial effusion determines the severity. In case of missing diameters, a label \textit{unknown severity} is assigned. A correct span includes both the diameter and location of the pericardial effusion.
\begin{itemize}
    \item Normal: $0$-$5$mm pericardial effusion, trace amounts
    \item Mildly abnormal: $5$-$10$mm pericardial effusion
    \item Moderately abnormal: $10$-$20$mm pericardial effusion
    \item Severely abnormal: $\ge$$20$mm pericardial effusion
    \item Unknown severity: Presence of pericardial effusion without corresponding severity
\end{itemize}

\subsection{Wall motion abnormalities}
A correct span includes both the type of wall motion abnormality (akinesia, dyskinesia, hypokinesia) and location (anterior, inferior, etc). 
\begin{itemize}
    \item Normal: Absence of wall motion abnormalities
    \item Present: Presence of wall motion abnormalities
\end{itemize}

\section{Additional file 3: SetFit comparison}
\begin{landscape}
\begin{table}[htbp]
\caption{Semantic performance of document classification methods}
\label{tab:doc_class_setfit_performance}
\centering
    \begin{tabular}{c|ccc|ccc}
                             &  \multicolumn{3}{c|}{\textbf{SetFit (RobBERT)}} & \multicolumn{3}{c}{\textbf{SetFit (BioLord)}} \\
\textbf{Characteristic}               & F1 & recall & precision & F1 & recall & precision  \\
         \hline
  Aortic regurgitation                & \textbf{0.93} (\textbf{0.86}) & \textbf{0.93} (\textbf{0.88}) & \textbf{0.93} (\textbf{0.84})    & 0.82 (0.66) & 0.82 (0.72) & 0.82 (0.66) \\
  Aortic stenosis                     & \textbf{0.91} (\textbf{0.82}) & \textbf{0.91} (\textbf{0.93}) & \textbf{0.91} (\textbf{0.75})    & 0.84 (0.66) & 0.84 (0.74) & 0.84 (0.60) \\
  Diastolic dysfunction               & \textbf{0.95} (\textbf{0.91}) & \textbf{0.95} (\textbf{0.97}) & \textbf{0.95} (\textbf{0.87})    & 0.93 (0.84) & 0.93 (0.87) & 0.93 (0.83) \\
  Left ventricular dilatation         & \textbf{0.95} (0.91) & 0.95 (0.95) & 0.95 (0.87)    & 0.92 (\textbf{0.95}) & 0.95 (0.95) & 0.95 (\textbf{0.89}) \\
  Left ventricular systolic dysfunction  & \textbf{0.95} (\textbf{0.91}) & \textbf{0.95} (\textbf{0.92}) & \textbf{0.95} (\textbf{0.89})    & 0.93 (0.90) & 0.93 (0.91) & 0.93 (0.89) \\
  Mitral regurgitation                & \textbf{0.94} (\textbf{0.88}) & \textbf{0.94} (\textbf{0.93}) & \textbf{0.94} (\textbf{0.85})    & 0.83 (0.74) & 0.83 (0.78) & 0.83 (0.71) \\
  Pericardial effusion                & \textbf{0.92} (\textbf{0.51}) & \textbf{0.92} (\textbf{0.60}) & \textbf{0.92} (\textbf{0.49})    & 0.88 (0.45) & 0.88 (0.51) & 0.88 (0.43) \\
  Right ventricular dilatation        & \textbf{0.92} (\textbf{0.81}) & \textbf{0.92} (\textbf{0.92}) & \textbf{0.92} (\textbf{0.74})    & 0.86 (0.71) & 0.86 (0.87) & 0.86 (0.62) \\
  Right ventricular systolic dysfunction & \textbf{0.94} (\textbf{0.89}) & \textbf{0.94} (\textbf{0.93}) & \textbf{0.94} (\textbf{0.85})    & 0.83 (0.71) & 0.83 (0.73) & 0.83 (0.71) \\
  Tricuspid regurgitation             & \textbf{0.92} (\textbf{0.83}) & \textbf{0.92} (\textbf{0.86}) & \textbf{0.92} (\textbf{0.80})    & 0.83 (0.65) & 0.83 (0.71) & 0.83 (0.63) \\
  Wall motion abnormalities           & \textbf{0.95} (\textbf{0.92}) & \textbf{0.95} (\textbf{0.93}) & \textbf{0.95} (\textbf{0.91})    & 0.94 (0.89) & 0.94 (0.92) & 0.94 (0.88) \\
    \end{tabular}
\footnotetext{Weighted and macro (in brackets) scores. The highest performance for each characteristic is denoted in bold.}
\end{table}

\begin{table}[htbp]
\caption{Semantic performance of document classification methods for simplified label scheme (\textit{No label}, \textit{Normal}, and \textit{Present})}
\label{tab:doc_class_setfit_performance_3labels}
\centering
    \begin{tabular}{c|ccc|ccc}
                             &  \multicolumn{3}{c|}{\textbf{SetFit (RobBERT)}} & \multicolumn{3}{c}{\textbf{SetFit (BioLord)}} \\
\textbf{Characteristic}               & F1 & recall & precision & F1 & recall & precision  \\
         \hline
  Aortic regurgitation                & \textbf{0.94 }(\textbf{0.93}) & \textbf{0.94} (\textbf{0.94}) & \textbf{0.94} (\textbf{0.91})     & 0.84 (0.81) & 0.84 (0.83) & 0.84 (0.80) \\
  Aortic stenosis                     & \textbf{0.91} (\textbf{0.86}) & \textbf{0.91} (\textbf{0.94)} & \textbf{0.91} (\textbf{0.82})     & 0.85 (0.77) & 0.85 (0.85) & 0.85 (0.73) \\
  Diastolic dysfunction               & \textbf{0.95} (\textbf{0.92}) & \textbf{0.95} (\textbf{0.97}) & \textbf{0.95} (\textbf{0.89})     & 0.93 (0.9)  & 0.93 (0.92) & 0.93 (0.88) \\
  Left ventricular dilatation         & 0.95 (\textbf{0.94}) & 0.95 (\textbf{0.96}) & 0.95 (\textbf{0.93})                     & 0.95 (0.93) & 0.95 (0.95) & 0.95 (0.92) \\
  Left ventricular systolic dysfunction  & \textbf{0.96} (\textbf{0.94}) & \textbf{0.96} (\textbf{0.95}) & \textbf{0.96} (\textbf{0.93})     & 0.93 (0.88) & 0.93 (0.9)  & 0.93 (0.86) \\
  Mitral regurgitation                & \textbf{0.94} (\textbf{0.94}) & \textbf{0.94} (\textbf{0.95}) & \textbf{0.94} (\textbf{0.93})     & 0.83 (0.82) & 0.83 (0.83) & 0.83 (0.81) \\
  Pericardial effusion                & \textbf{0.95} (\textbf{0.85}) & \textbf{0.95} (\textbf{0.93}) & \textbf{0.95} (\textbf{0.79})     & 0.87 (0.68) & 0.87 (0.79) & 0.87 (0.63) \\
  Right ventricular dilatation        & \textbf{0.91} (\textbf{0.86}) & \textbf{0.91} (\textbf{0.91}) & \textbf{0.91} (\textbf{0.83})     & 0.88 (0.81) & 0.88 (0.87) & 0.88 (0.77) \\
  Right ventricular systolic dysfunction & \textbf{0.93} (\textbf{0.91}) & \textbf{0.93} (\textbf{0.94}) & \textbf{0.93} (\textbf{0.90})     & 0.85 (0.83) & 0.85 (0.86) & 0.85 (0.81) \\
  Tricuspid regurgitation             & \textbf{0.93} (\textbf{0.91}) & \textbf{0.93} (\textbf{0.93}) & \textbf{0.93} (\textbf{0.89})     & 0.85 (0.79) & 0.85 (0.82) & 0.85 (0.77) \\
  Wall motion abnormalities           & \textbf{0.95} (\textbf{0.92}) & \textbf{0.95} (\textbf{0.93}) & \textbf{0.95} (\textbf{0.91})     & 0.94 (0.90) & 0.94 (0.92) & 0.94 (0.88)  \\
    \end{tabular}
\footnotetext{Weighted and macro (in brackets) scores. The highest performance for each characteristic is denoted in bold.}
\end{table}
\end{landscape}
\end{document}